\documentclass[onecolumn,11pt,draftcls]{IEEEtran}
\usepackage{graphics,multirow,amsmath,amssymb,textcomp,subfigure,multirow,xspace,arydshln,cite}
\usepackage[pdftex]{graphicx}
\DeclareGraphicsExtensions{.eps,.pdf,.jpeg,.png}
\usepackage{enumerate}
\usepackage{amsmath, amsfonts}
\usepackage{layout}
\usepackage{color}
\usepackage[ruled,vlined,boxed,linesnumbered]{algorithm2e}
\usepackage[pagebackref=true,breaklinks=true,letterpaper=true,colorlinks,bookmarks=false,linkcolor=black,citecolor=black,urlcolor=blue]{hyperref}
\LinesNumberedHidden
\DontPrintSemicolon
\makeatletter
\DeclareRobustCommand\onedot{\futurelet\@let@token\@onedot}
\def\@onedot{\ifx\@let@token.\else.\null\fi\xspace}

\makeatother

\begin{document}

\title{Fast Multi-Layer Laplacian Enhancement}
\author{Hossein Talebi and Peyman Milanfar\\
\thanks{H. Talebi and P. Milanfar are with Google Inc., Mountain View, USA, Email: \{htalebi, milanfar\}@google.com.}
}

\maketitle
\vspace{-2.5cm}

\begin{figure*}[!h]
\vspace{-0 mm}
\begin{center}
\subfigure{
\includegraphics*[viewport=270 150 670 530,scale=0.28]{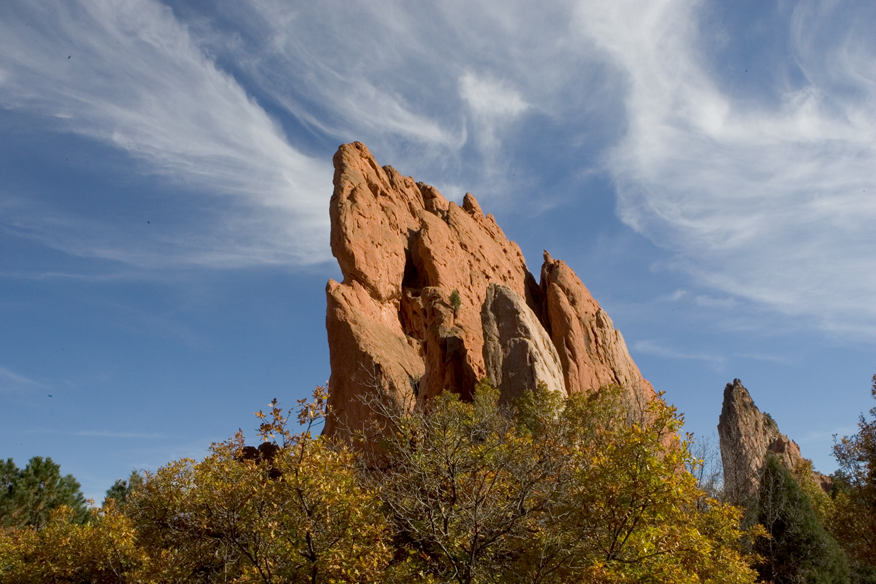}}
\subfigure{
\includegraphics*[viewport=270 150 670 530,scale=0.28]{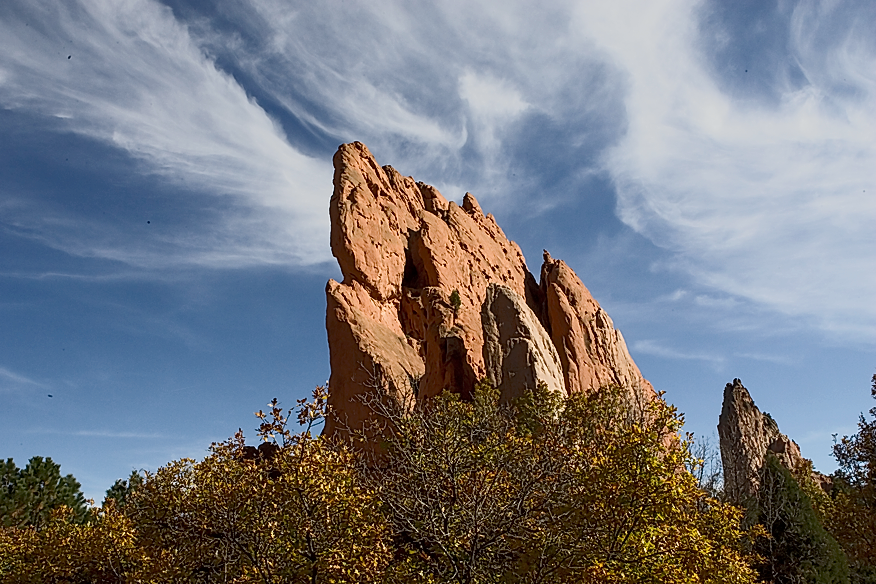}}
\subfigure{
\includegraphics*[viewport=490 720 850 1145,scale=0.25]{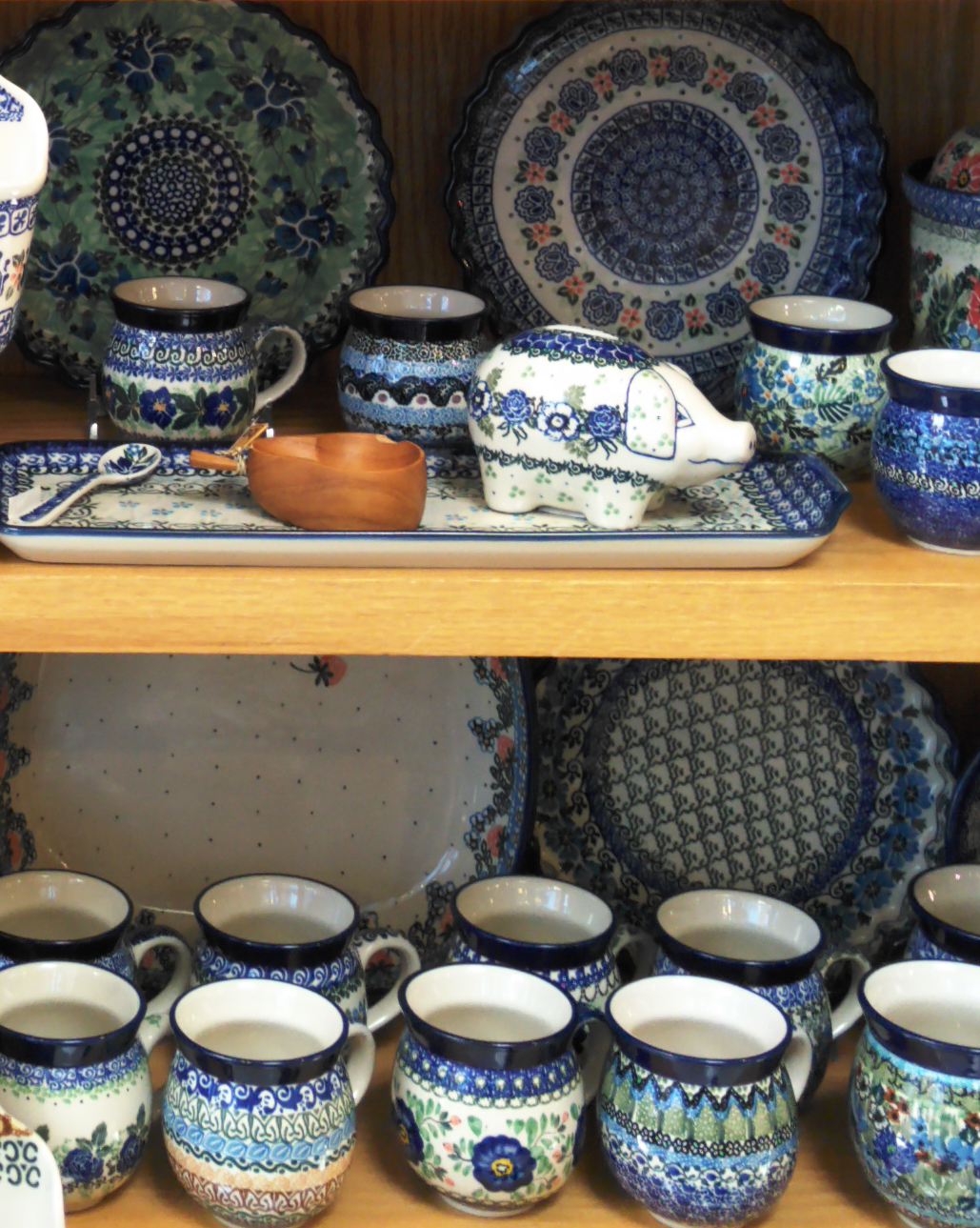}}
\subfigure{
\includegraphics*[viewport=490 720 850 1145,scale=0.25]{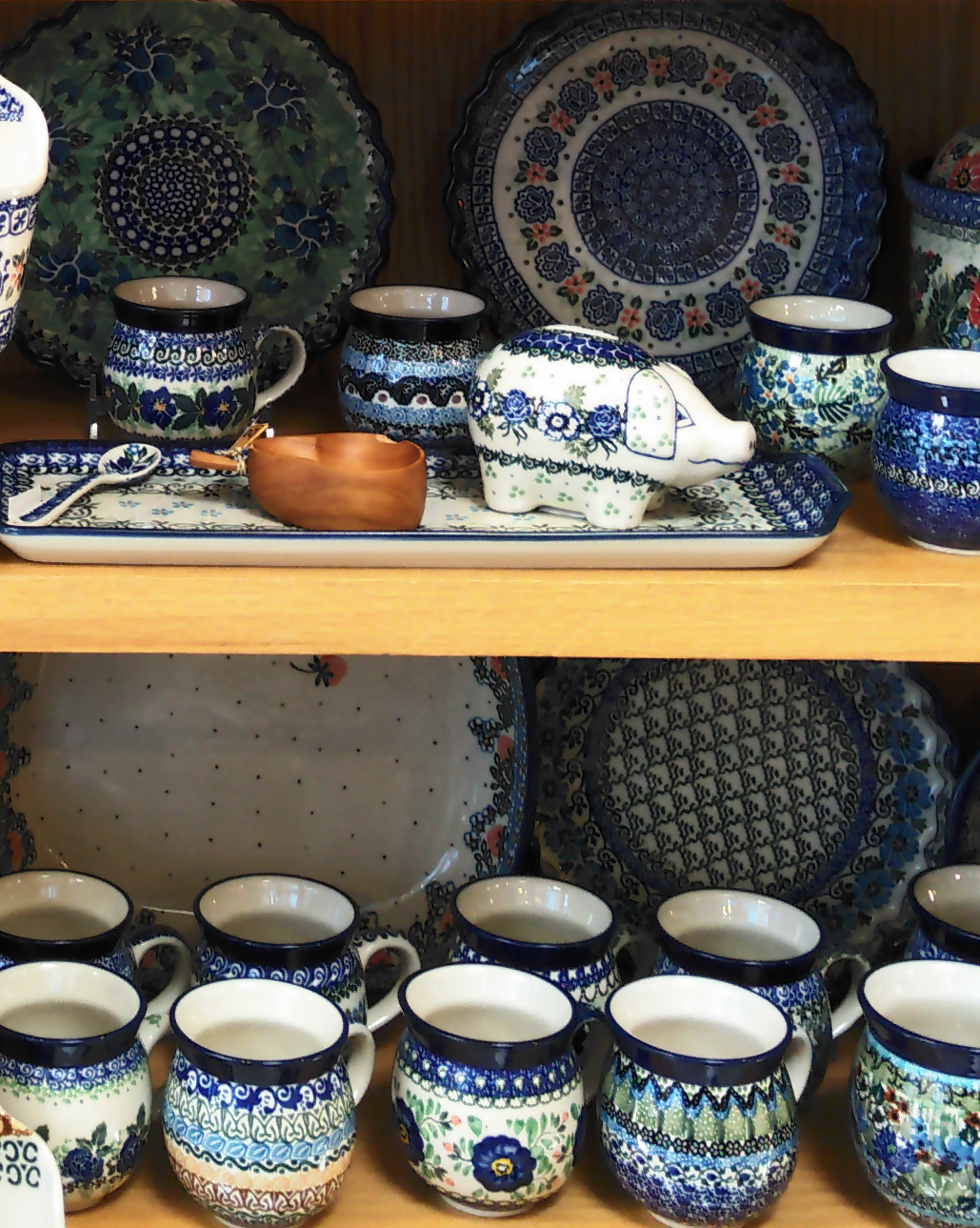} }
\subfigure{
\includegraphics*[viewport=310 300 630 600,scale=0.4]{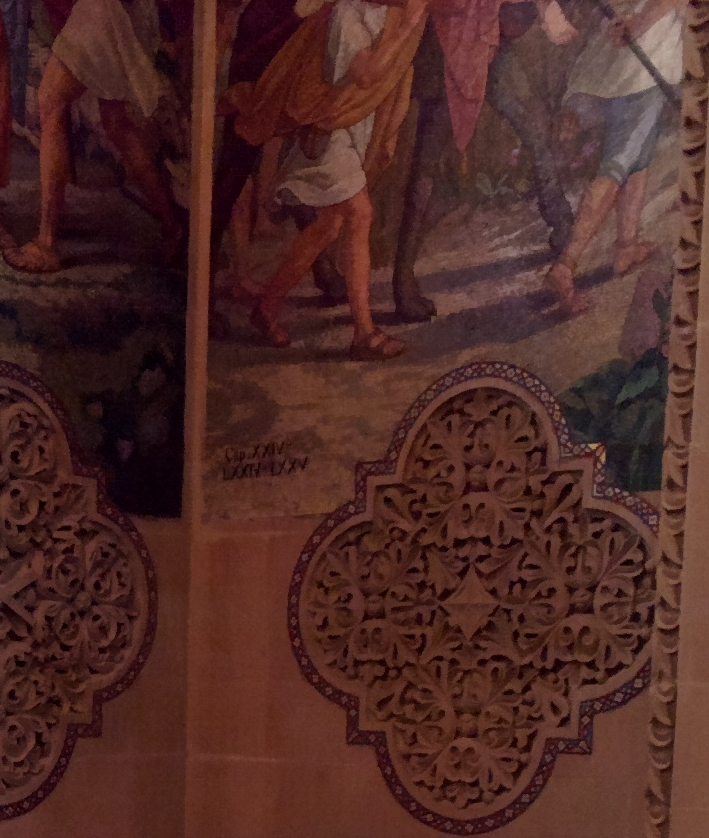}}
\subfigure{
\includegraphics*[viewport=310 300 630 600,scale=0.4]{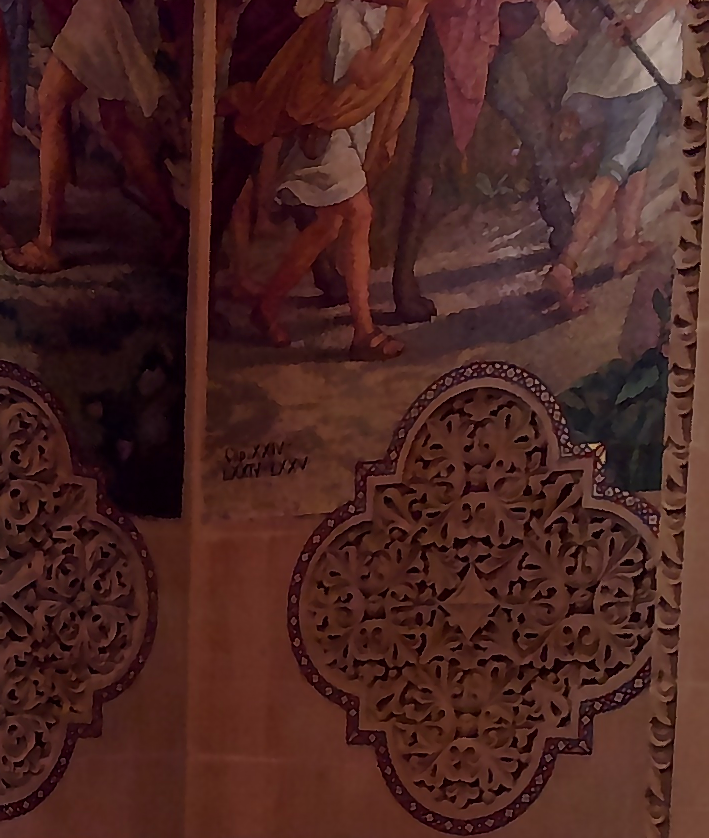}}
\subfigure{
\includegraphics*[scale=0.32]{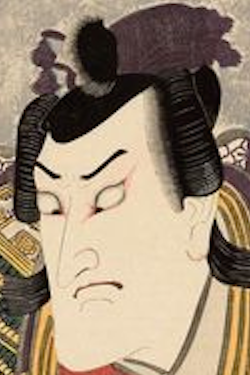}}
\subfigure{
\includegraphics*[scale=0.32]{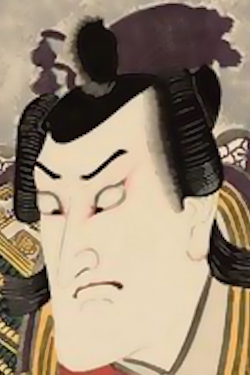}}
\end{center}
\vspace{-7 mm}
{\caption{Examples of detail enhancement using our method. Each image pair shows the input (left) and enhanced image (right). Top row: Application of our method for sharpening and local contrast enhancement. Bottom row: Application of our method for simultaneous artifact/noise removal and sharpening. \label{fig:logo}}}
\vspace{-5 mm}
\end{figure*}

\begin{abstract}
\vspace{-3 mm}
A novel, fast and practical way of enhancing images is introduced in this paper. Our approach builds on Laplacian operators of well-known edge-aware kernels, such as bilateral and nonlocal means, and extends these filter's capabilities to perform more effective and fast image smoothing, sharpening and tone manipulation. We propose an approximation of the Laplacian, which does not require normalization of the kernel weights. Multiple Laplacians of the affinity weights endow our method with progressive detail decomposition of the input image from fine to coarse scale. These image components are blended by a structure mask, which avoids noise/artifact magnification or detail loss in the output image. Contributions of the proposed method to existing image editing tools are: (1) Low computational and memory requirements, making it appropriate for mobile device implementations (e.g. as a finish step in a camera pipeline), (2) A range of filtering applications from detail enhancement to denoising with only a few control parameters, enabling the user to apply a combination of various (and even opposite) filtering effects.

\end{abstract}
\vspace{-5 mm}
\begin{IEEEkeywords}
\vspace{-3 mm}
Image Enhancement, Image Editing, Image sharpening, Tone mapping, Image smoothing
\end{IEEEkeywords}

\section{Introduction}
\label{sec:Introduction}
\vspace{0 mm}

Recently, edge-preserving image operators have been widely used in image enhancement applications. These filters allow separate processing of texture and piecewise smooth components of the image. Given that the main structure (edges) of the images are preserved by these edge-aware filters, applying an appropriate nonlinearity on the texture component results in local contrast enhancement and tonal adjustment \cite{GIF, WGIF,FLLF,BilateralTexture, SPI_Smoothing, Farbman2008, Subr2009, Sylvain2011, Farbman2010}. However, when using these methods, the default assumption is that undesired perturbations, such as noise or compression artifacts are removed beforehand. In practical imaging scenarios, boosting a detail image layer can result in noise and artifact magnification, limiting applications of the existing detail enhancement algorithms. This issue is mitigated in our proposed method by employing a new blending strategy, which smoothes regions containing noise while sharpening significant image details. Our experiments demonstrate that the proposed method can be effective in improving details and local contrast of images, whilst efficiently handling mildly degraded cases (examples of the proposed method's applications are illustrated in Fig. \ref{fig:logo}). The existing relevant literature is reviewed next.

\subsection{Related Work}
\label{sec:RelatedWork}
\vspace{0 mm}

Linear unsharp masking (UM) is perhaps the simplest algorithm for enhancing the edge and detail information of an image. Linear UM is a high-pass filter, which sharpens high frequency content of images, yet magnifies noise and produces undesirable distortions, such as halo artifacts. Polesel et al. \cite{AUM} proposed adaptive unsharp mask to improve on the classic UM. This method measures local image gradient to adaptively apply the UM filter on details, and leave flat regions unchanged. Constrained unsharp mask \cite{C_UM} is another alternative, which combines a denoised and a sharpened version of the input image. Overall, the linear smoothing filter employed at the core of these methods can restrict their performances.  Replacing the linear operator with a data-dependent (non-linear) smoother diminishes this issue.

The Bilateral filter is possibly the most widely used edge-aware filter in image processing and computer graphics \cite{tomasi}. Similarity of neighbor pixels is measured by bilateral range filter, avoiding averaging across principal edges. Durand and Dorsey \cite{Durand2002} exploit application of the bilateral filter in contrast reduction of high dynamic range images. A multi-scale implementation of the bilateral filter for progressive detail extraction is explored in \cite{Fattal2007}. Variations of bilateral filter can also be used for sharpening \cite{ABF}, creating cartoon effects \cite{BilateralTexture}, image editing \cite{Chen2007} and abstraction \cite{VidAbs}. Although bilateral filter outperforms linear smoothers, it still lacks robustness in some applications such as denoising.

Nonlocal means filter (NLM) works similarly to the bilateral kernel, except that photometric similarity of neighboring pixels is determined by measuring patch distances \cite{buades_nonlocal, nlm2, OSA}. NLM weights better handle noise and other image distortions compared to bilateral kernel, yet offer competitive smoothing properties. Choudhury et al. \cite{Choudhury} propose a multi-scale sharpness enhancement scheme based on NLM weights. A noise suppression step is performed first, and then different detail layers are extracted by recomputing NLM weights several times with various smoothing parameters. However, multiple realizations of the NLM filter impose a high computational complexity on this algorithm. NLM affinity weights were also used for various image editing tasks, such as tone manipulation and edit propagation in \cite{NLEditing}. In that work, the global affinity matrix is approximated by its leading eigenvectors, enabling different filtering effects by polynomial mapping of the filter eigenvalues. More recently, differences of NLM smoothers are used to sharpen mildly blurred images \cite{AminSharpener}. Overall, global filtering parameters, filter weight computation and memory storage may limit application of these methods.

More nonlinear filters have been introduced in the past few years. A progressive coarsening operator based on a weighted least square optimization is proposed in \cite{Farbman2008}. Subr et al. \cite{Subr2009} introduced a new image decomposition method by smoothing large image oscillations and preserving edges. Gaussian pyramids are also used for edge-aware filtering in local Laplacian framework \cite{Sylvain2011,FLLF}, where each detail layer is mapped by a specific function, resulting in tonal enhancement of the reconstructed image. The domain transform paradigm proposed by Gastal et al. \cite{DomainTransform} formulates the nonlinear image smoothing as a few iterations of one dimensional filtering. The main edges are detected by image gradient and preserved in the filtering process. Another gradient-based smoother is introduced in \cite{SPI_Smoothing} where image structure and texture are distinguished by means of local covariance. Edge preserving operators are also practically viable by guided image filtering \cite{GIF, WGIF}. Image smoothing while constraining the number of non-zero gradients is another edge-aware filtering technique \cite{l0_smoothing}. This approach removes low-amplitude structures by progressively reducing the number of non-zero gradients. Similar to the framework in \cite{l0_smoothing}, an $L_1$ energy minimization method for image smoothing is proposed in \cite{l1_smoothing}. The energy cost includes local variations and global sparsity terms, and minimizing it results in flattening details. Our intention in this paper is not to introduce yet another nonlinear smoother. In fact, the base smoothing filter upon which the rest of the presented framework is constructed can be any of the existing filters mentioned above.

In addition to edge preserving filters, other contrast enhancement techniques based on the retinex theory \cite{retinex} have been developed in the past few years \cite{multi_retinex_1997, retinex_2003}. Retinex theory explains how humans can see colors consistently in spite of difference in light levels. Inspired by this theorem, several enhancement algorithms have been proposed recently \cite{retinex_2003, retinex_2010, multi_retinex}. Although these techniques are quite efficient and produce compelling results, noise magnification while brightening dark pixels remains challenging.
 
 \begin{figure*}[!t]
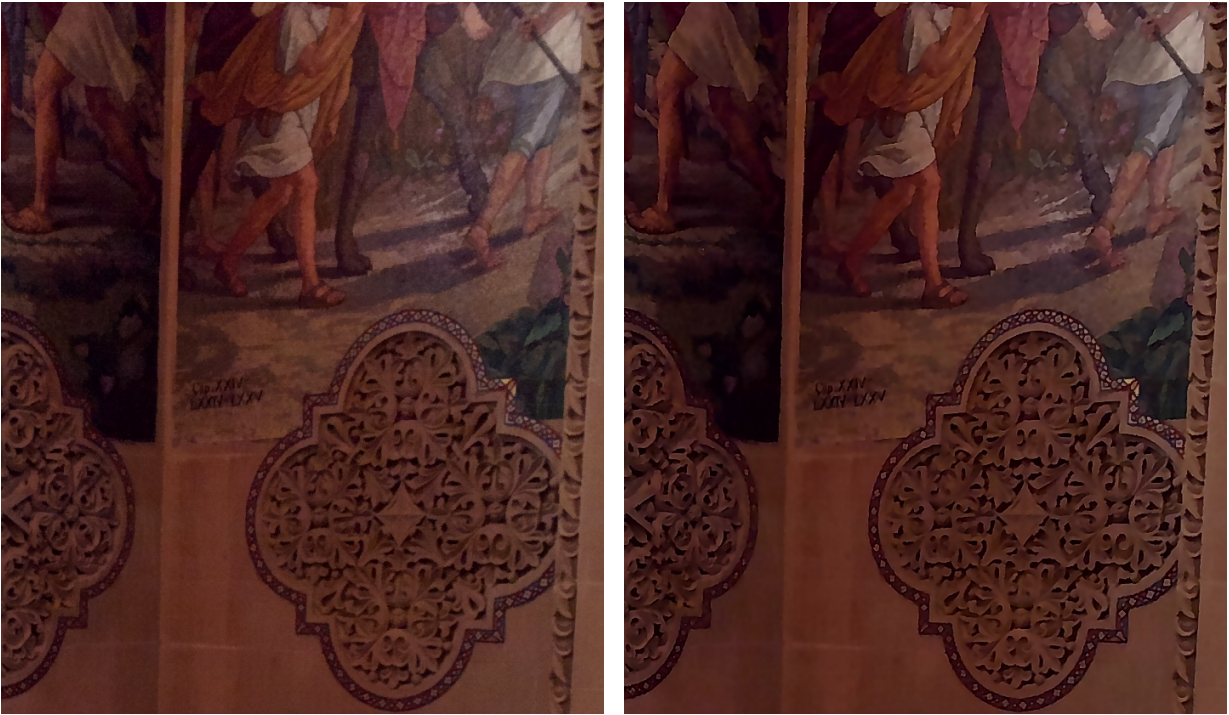

\vspace{-0 mm}
\begin{center}
\subfigure{
\includegraphics*[scale=0.32]{./figures/church.png}}
\subfigure{
\includegraphics*[scale=0.32]{./figures/church_enhance.png}}
\end{center}
\vspace{-7 mm}
{\caption{ Application of our method for simultaneous artifact/noise removal and sharpening. The input image is of size $1856\times 2528$ and average running time for producing this result on an Intel Xeon CPU @ 3.5 GHz is about $0. 2$ seconds.\label{fig:church}}}
\vspace{-5 mm}
\end{figure*}

\subsection{Contributions}
\label{sec:Contributions}
\vspace{0 mm}

The Laplacian operator of the local affinity matrix is at the core of our algorithm. The Laplacian operators can be computed for any smoothing operator, yet we develop our method based on the NLM kernel which is quite resilient to noise. Contributions of this work to the current image enhancement literature are:  
\begin{itemize}
  \item \textit{A novel filtering approach using normalization-free filters}: Affinity weights are conventionally normalized and applied on the image to preserve the local brightness. In this paper we propose an efficient approximation of the normalized affinities to provide a computationally simplified \textit{un-normalized} filtering paradigm.
  \item \textit{Detail manipulation in the presence of mild image distortions}: Instead of applying noise/artifact suppression as a pre-filtering stage (which imposes extra complexity to the framework and may remove image details), our approach naturally handles these degradations. More specifically, fine detail boosting is replaced by smoothing when dealing with noisy regions. Fig. \ref{fig:church}  demonstrates an example of simultaneous smoothing and sharpening using our proposed method. 
  \item \textit{Substantial complexity reduction of nonlinear multi-scale decomposition}: We propose a simple, yet effective way to compute the multi-scale detail decomposition by approximating affinity weights. Typical nonlinear multi-scale decomposition relies on successive computation of the filter weights on the input image. Given the exponential affinities of the NLM and bilateral kernels, we precompute the image-dependent filter weights only once and produce different versions of the filter by simple direct product of the weights. Significant speed up is observed by this strategy.
\end{itemize}

\begin{figure*}[!t]
\vspace{-0 mm}
\begin{center}
\subfigure[\scriptsize Input]{
\includegraphics*[scale=0.23]{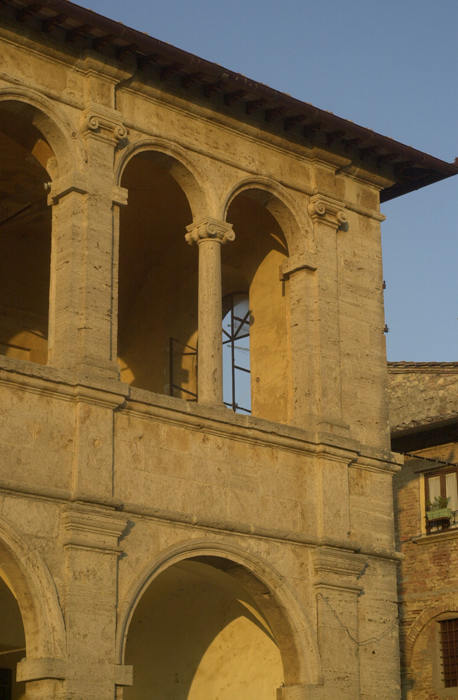}}
\subfigure[\scriptsize Smoothed]{
\includegraphics*[scale=0.23]{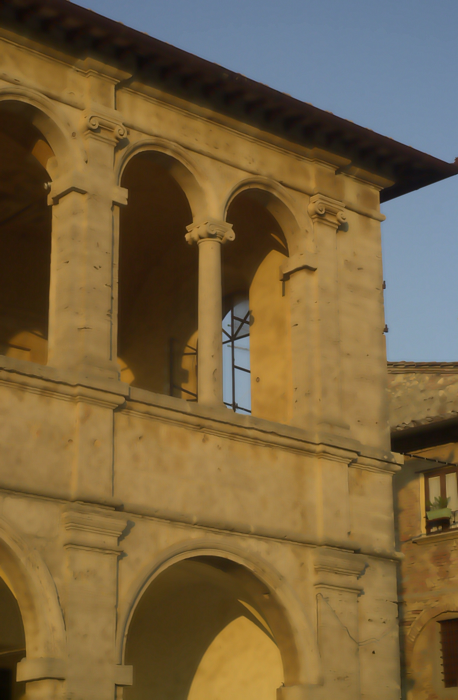} }
\subfigure[\scriptsize Enhanced (effect 1)]{
\includegraphics*[scale=0.23]{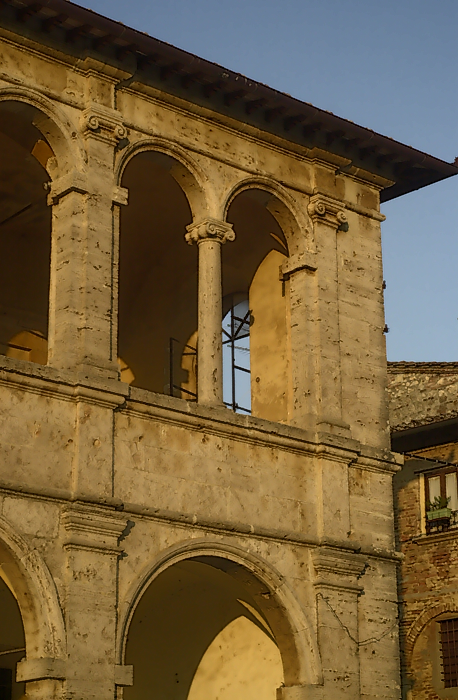} }
\subfigure[\scriptsize Enhanced (effect 2)]{
\includegraphics*[scale=0.23]{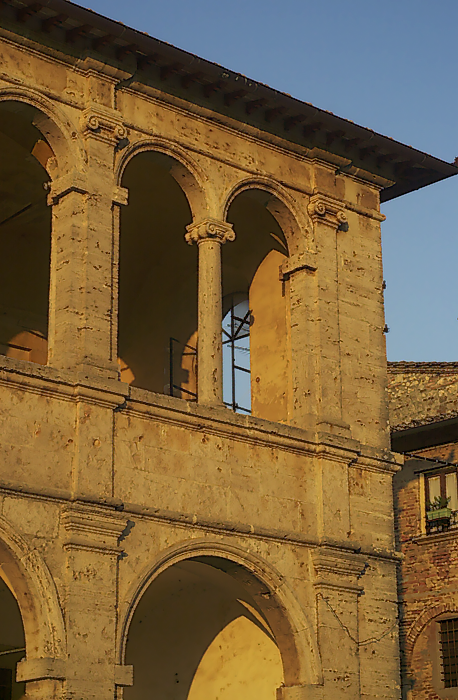} }
\end{center}
\vspace{-5 mm}
{\caption{Example of our method applied for (b) detail smoothing, (c) detail enhancement in luma channel, and (d) detail enhancement in RGB channels. The input image is of size $700\times 458$ and average running time for producing effect 1 is about $0. 015$ second. \label{fig:house}}}
\vspace{-0 mm}
\end{figure*}

The proposed method allows real-time detail manipulation and enhancement such as examples shown in Fig. \ref{fig:house}. The rest of the paper is organized as follows. In Section \ref{sec:proposed}, a detailed explanation of the proposed method is described. Next, in Section \ref{sec:results}, applications of our algorithm are exemplified. We also provide details of our implementation along with running time comparisons. Finally, this paper is concluded in Section \ref{sec:conclusion}.

\section{Proposed scheme}
\label{sec:proposed}
\vspace{0 mm}

\begin{figure*}[!t]
\vspace{-0 mm}
\begin{center}
\subfigure{
\includegraphics*[viewport=1 100 750 370,scale=0.65]{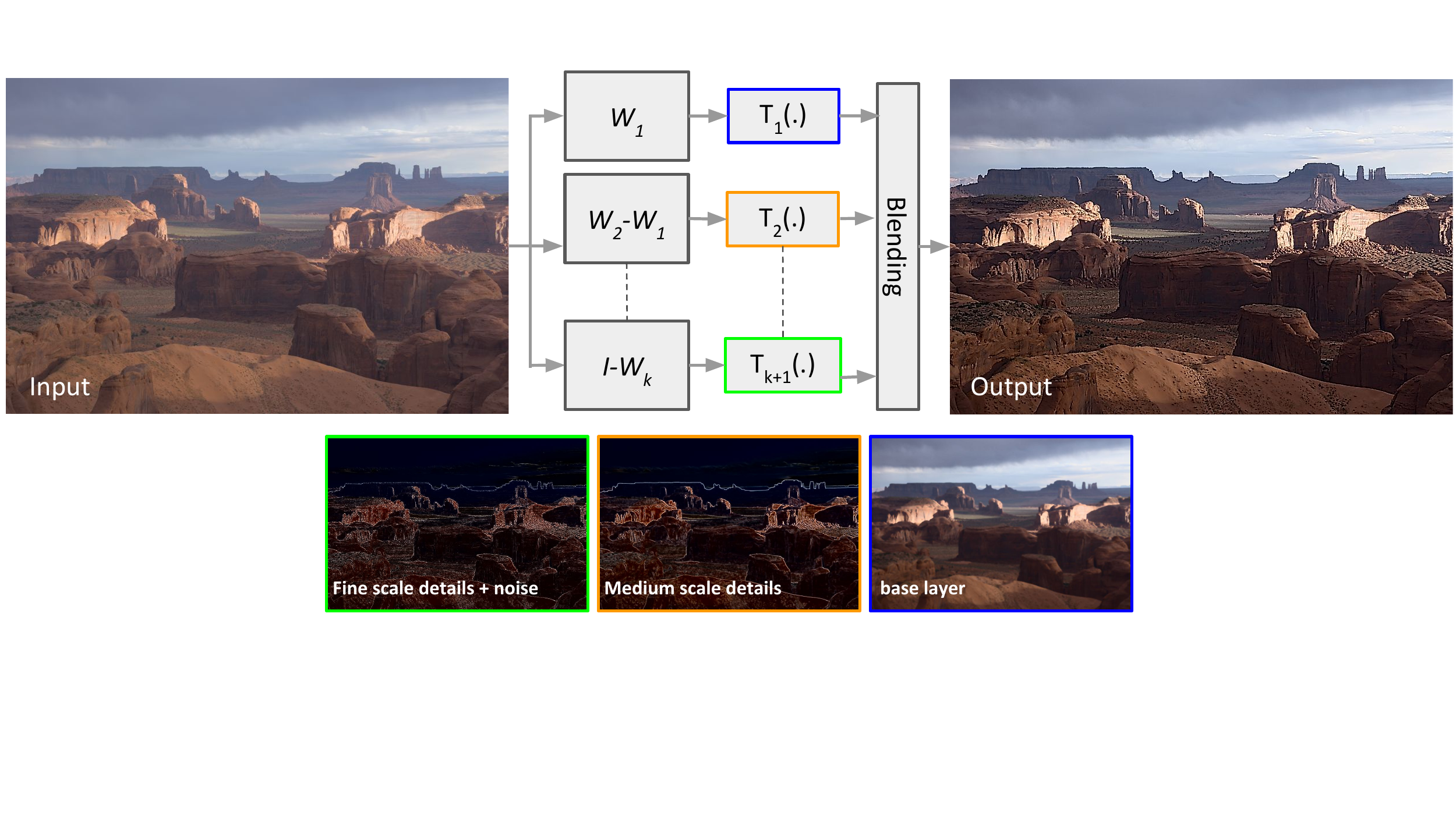}}
\end{center}
\vspace{-7 mm}
{\caption{The proposed pipeline: The input image is fed to multiple affinity based Laplacians obtained from NLM kernel ($\textbf{W}_l$) to produce various detail layers. The filter operators are $\{\textbf{W}_1, \textbf{W}_2-\textbf{W}_1, ..., \textbf{W}_{k}-\textbf{W}_{k-1}, \textbf{I} - \textbf{W}_k\}$ and the smoothest image layer is obtained from $\textbf{W}_1$ (Given the smoothing parameter of $\textbf{W}_l$ as $h_l$, for every $1 \leq l < k$ we have $h_l > h_{l+1}$). Detail layers are mapped by nonlinear s-curves ($T_l(.)$) and blended by a structure mask to produce the enhanced image. \label{fig:pipeline}}}
\vspace{-0 mm}
\end{figure*}

Our enhancement algorithm is broadly illustrated in Fig. \ref{fig:pipeline}. The input image is filtered by different affinity-based operators (built on the NLM weights \cite{buades_nonlocal}) to produce the image detail layers. Each layer is mapped by a nonlinear function to boost or suppress the associated detail and then, the manipulated layers are blended through a structure mask. The proposed scheme has parameters of the mapping functions as its filtering knobs, which control the filter's behavior by altering it from smoothing to sharpening and from tone enhancement to tone compression. Our processing is principally in the YUV domain, by filtering luma and leaving chroma unaltered. For completeness, we also illustrate our method for filtering RGB color channels separately (Fig. \ref{fig:house}). In this section, first the affinity filters are reviewed and then the normalization-free filter weights are discussed. We elaborate on the details of the proposed filtering scheme, and finally our mapping functions and blending strategy are discussed.

\subsection{Nonlinear edge-aware filters}
\label{sec:nlf}
\vspace{0 mm}

The general construction of an edge-aware filter begins by specifying a symmetric positive semi-definite (PSD) kernel $k_{ij} \geq 0$ that measures the similarity, or affinity, between individual or groups of pixels. This affinity can be measured as a function of both the distance between the spatial variables (denoted by $\textbf{x}$), but more importantly, also using the gray or color value (denoted by $\textbf{y}$). While the results of this paper extend to any filter with a PSD kernel, some popular examples commonly used in the image processing, computer vision, and graphics literature are as follows:

\paragraph{Bilateral (BL) \cite{tomasi,elad1}} This filter takes into account both the spatial \textit{and} value distances between two pixels, generally in a separable fashion. For BL we have:
\begin{equation}
\label{eqn:bilat}
k_{ij} = \exp \left(\frac{-\|\textbf{x}_i-\textbf{x}_j\|^2}{h_x}\right)\,\exp\left(\frac{-(y_i-y_j)^2}{h_y}\right) 
\end{equation}
As seen in the overall exponent, the similarity metric here is a weighted Euclidean distance between the concatenated vectors $(\textbf{x}_i,y_i)$ and $(\textbf{x}_j,y_j)$. 

\paragraph{Nonlocal Means (NLM)  \cite{buades_nonlocal,OSA}} The NLM kernel is a generalization of the bilateral kernel in which the value distance term is measured patch-wise instead of point-wise:
\begin{equation}
\label{eqn:nlm}
k_{ij} = \exp\left(\frac{-\|\textbf{x}_i-\textbf{x}_j\|^2}{h_x}\right)\,\exp\left(\frac{-\|\textbf{y}_i-\textbf{y}_j\|^2}{h_y}\right),
\end{equation}
where $\textbf{y}_i$ and $\textbf{y}_j$ refer now to {\em subsets} of samples (i.e. patches) in $\textbf{y}$.


These affinities are not used directly to filter the images, but instead in order to maintain the local average brightness, they are normalized so that the resulting weights pointing to each pixel sum to one. More specifically,
\begin{equation}
\label{eqn:normweights}
w_{ij} = \frac{k_{ij}}{\sum_{j=1}^n k_{ij}},
\end{equation}
\noindent where each element of the filtered signal $\textbf{z}$ is then given by
\begin{equation}
z_i = \sum_{j=1}^n w_{ij} \; y_j.
\end{equation}
\noindent It is worth noting that the denominator in (\ref{eqn:normweights}) can be computed by simply applying the filter (without normalization) to an image of all $1$'s.
 
In matrix notation, the collection of the weights used to produce the $i$-th output pixel is the vector $\left[w_{i1},\;\cdots,\; w_{in} \right]$; and this can in turn be placed as the $i$-th row of a filter matrix $\textbf{W}$ so that 
\begin{equation}
\textbf{z} = \textbf{W} \textbf{y}.
\end{equation}
We note again that due to the normalization of the weights, the rows of the matrix $\textbf{W}$ sum to one, That is, for each $1 \leq i \leq n$, 
\begin{equation}
\sum_{j=1}^n w_{ij}=1.
\end{equation}
Viewed another way, the filter matrix $\textbf{W}$ is a {\em normalized} version of the symmetric positive definite affinity matrix $\textbf{K}$ constructed from the {\em un-normalized} affinities $k_{ij}$, $1\leq i,\:j, \leq n$. As a result, $\textbf{W}$ can be written as a product of two matrices 
\begin{equation}
\label{eqn:diagonalform}
\textbf{W} = \textbf{D}^{-1} \textbf{K},
\end{equation}
where $\textbf{D}$ is a diagonal matrix with diagonal elements $\left[\textbf{D}\right]_{ii} = \sum_{j=1}^n k_{ij} = d_i$.

\begin{figure*}[!t]
\vspace{-0 mm}
\begin{center}
\subfigure{
\includegraphics*[viewport=1 30 720 550,scale=0.28]{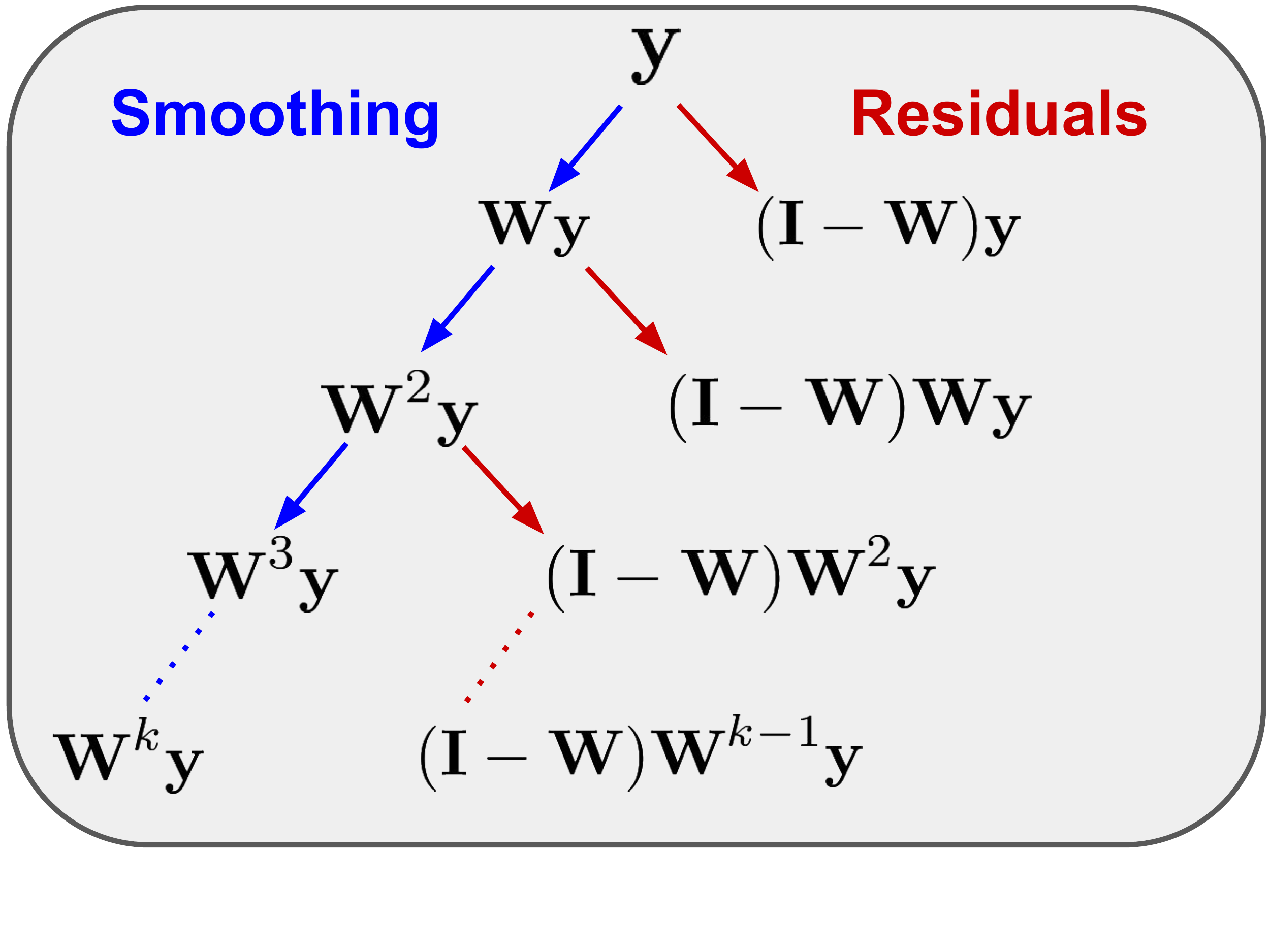}}
\end{center}
\vspace{-7 mm}
{\caption{The proposed image decomposition framework. Input image $\textbf{y}$ is decomposed into smoothed $\textbf{W}\textbf{y}$ and residual $(\textbf{I}-\textbf{W})\textbf{y}$ components. $k$ iterations of this process on the smoothed image leads to $k$ residual layers. \label{fig:image_dec}}}
\vspace{-0 mm}
\end{figure*}

\subsection{Motivation}
\label{sec:mtv}
\vspace{0 mm}

In our multiscale scheme, the filtered image $\textbf{z}$ is expressed as a linear combination of $k$ detail layers and one smooth layer:
\begin{equation}
\textbf{z} = \beta_1 \: \textbf{y}_{smooth} + \beta_2 \: \textbf{y}_{detail_1} + \cdots + \beta_{k+1} \: \textbf{y}_{detail_k}
\end{equation} 
where $\beta_i$ denotes the boosting or shrinking factor associated with each layer. A nonlinear filter $\textbf{W}$ can be used to realize this decomposition scheme as:
\begin{equation}
\label{eqn:multi}
\textbf{z} = \beta_1 \textbf{W}^k\textbf{y} + \beta_2 (\textbf{I}-\textbf{W})\textbf{W}^{k-1}\textbf{y} + \cdots + \beta_k(\textbf{I}-\textbf{W})\textbf{W}\textbf{y} + \beta_{k+1}(\textbf{I} - \textbf{W})\textbf{y}
\end{equation} 
in which $\textbf{W}^k\textbf{y}$ represents the smooth layer obtained from $k$ diffusion iterations of $\textbf{W}$ filter. The remaining terms consist of $k-1$ ``band-pass" and one ``high-pass" filters that decompose the input image into various detail layers (This image decomposition scheme is shown in Fig. \ref{fig:image_dec}). As shown in \cite{NLEditing}, the multiscale filtering in (\ref{eqn:multi}) corresponds to a polynomial mapping of the filter eigenvalues. Although this interpretation provides a flexible global framework for affinity based filtering in the spectral domain, complexity of the eigen-decomposition approximation remains relatively high.

Our current paper's motivation is to expedite the local computation of the diffusion filtering in (\ref{eqn:multi}). More explicitly, given the filter $\textbf{W}$ of size $n\times n$, the expensive matrix multiplication of the diffusion process in computing $\textbf{W}^k$ ($\mathcal{O}(kn^{3})$) is replaced by recomputation of the affinity weights using a larger smoothing parameter $h_k$. As it will be addressed in this work, we only compute the affinity weights once and reuse the filter weights to efficiently reevaluate the affinity kernel weights. This leads to a quadratic filter computation complexity of $\mathcal{O}(kn^2)$. In what follows, the normalization-free filter is discussed first, and then, our multiscale enhancement scheme is described in more details.  

\subsection{The Normalization-free Filter}
\label{sec:nff}
\vspace{0 mm}

To avoid the normalization in (\ref{eqn:diagonalform}), we will replace the filter $\textbf{W}$ with an approximation $\widehat{\textbf{W}}$ that only involves $\textbf{D}$ rather than its inverse. More specifically,
\begin{equation} 
\label{eqn:approxfilt}
\widehat{\textbf{W}} = \textbf{I} +   \alpha (\textbf{K} - \textbf{D}). 
\end{equation} 
Why is this a good idea? In what follows, we will motivate and derive this approximation from first principles, while also providing an analytically sound and numerically tractable choice for the scalar $\alpha>0$ that gives the best approximation to $\textbf{W}$ in the least-squares sense. Before doing so, it is worth noting some of the key properties and advantages of this approximate filter which are evident from the above expression (\ref{eqn:approxfilt}). 
\begin{itemize}
\item Regardless of the value of $\alpha$, the rows of $\widehat{\textbf{W}}$ always sum to one. That is, like its counterpart $\textbf{W}$ constructed with $\textbf{D}^{-1}$, the approximation $\widehat{\textbf{W}}$, constructed with only $\textbf{D}$, is automatically normalized. This can be easily seen by multiplying $\widehat{\textbf{W}}$ on the right by a vector of ones, and observing that it returns the same vector back regardless of $\alpha$.  
\item While the filter $\textbf{W}$ is not symmetric due to the multiplicative normalization (see Eq. \ref{eqn:diagonalform}), the approximate $\widehat{\textbf{W}}$ is always symmetric, again regardless of $\alpha$. The advantages of having a symmetric filter matrix are many, as documented in the recent work \cite{symmetrizing}. 
\item The PSD affinity matrix $\textbf{K}$ is typically also non-negative {\em valued}, leading to filter weights in $\textbf{W}$ which are also in turn non-negative valued. The elements in $\widehat{\textbf{W}}$ however, can be negative valued due to the term $\textbf{K} - \textbf{D}$. This means that the behavior of the approximate filter may differ from its reference value, and must be carefully studied and controlled. We will do this next. 
\end{itemize}

To derive the approximation, we first note that the standard filter can be written as: 

\begin{equation} 
\label{eq:newform}
\textbf{W} = \textbf{I} + \textbf{D}^{-1}(\textbf{K}-\textbf{D})  	 
\end{equation}

\begin{figure*}[!t]
\begin{center}
 \includegraphics[width=0.4\textwidth]{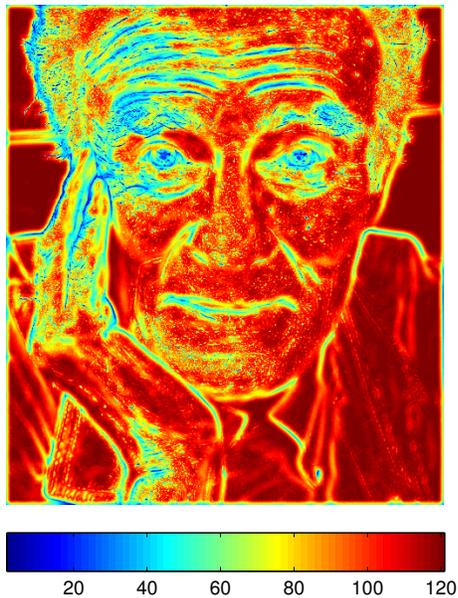}	
\end{center}
\vspace{-7 mm}
 \caption{Values of $d_i$ for the old man photo. Large values shown in red indicate pixels that have many ``nearest neighbors'' in the metric implied by the bilateral kernel. Weights were computed over 11$\times$11 windows (i.e. $m=121$).}
\label{fig:OldmanDs}
\end{figure*}

\begin{figure*}[ht]
\begin{center}
 \includegraphics*[viewport=1 1 1100 350,scale=0.4]{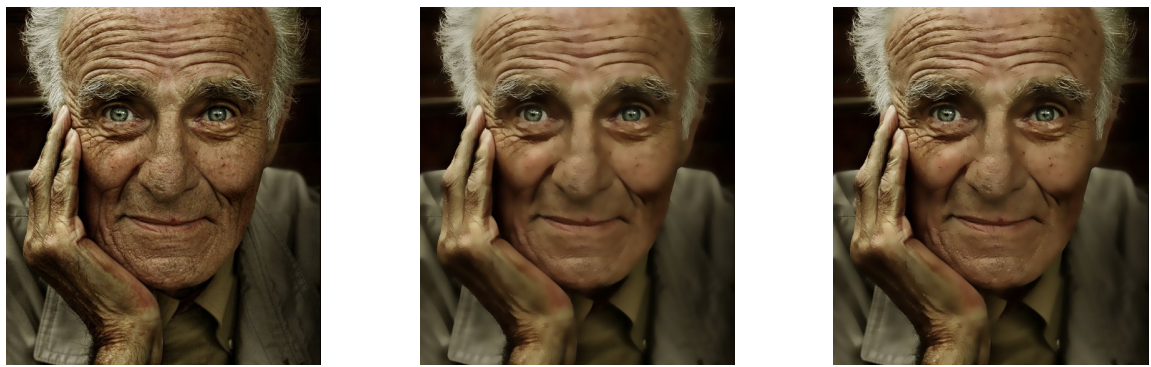}
\end{center}
\vspace{-7 mm}
 \caption{(Left) Input $\textbf{y}$; (Center) exact BL filter $\textbf{W}\textbf{z}$ , and (right) approximate BL filter $\widehat{\textbf{W}}\textbf{z}$.}
\label{fig:Oldman}
\end{figure*}

Comparing this form to the one presented earlier in (\ref{eqn:approxfilt}), we note that the approximation is replacing the matrix inverse (on the right hand side) with a scalar multiple of the identity:
\begin{equation}
\textbf{D}^{-1} \approx \alpha \textbf{I}
\end{equation} 
As an illustration, an image containing the normalization terms $d_i$ (which comprise the diagonal elements of $\textbf{D}$) for the photo in Fig. \ref{fig:Oldman}, are shown in Fig. \ref{fig:OldmanDs}. The proposal, as we elaborate below, is to replace these many normalization constants in (\ref{eq:newform}) with a single constant.

The justification for this approximation is a Taylor series in terms of $\textbf{D}$ for the filter matrix. In particular, let's consider the first few terms in the series around a nominal $\textbf{D}_0$:
\begin{equation}
\label{taylor}
\textbf{D}^{-1} \textbf{K} \approx \textbf{I} + \textbf{D}_0^{-1}(\textbf{K}-\textbf{D}) -\textbf{D}_0^{-2}(\textbf{D} - \textbf{D}_0)(\textbf{K} -\textbf{D})
\end{equation}

The series expresses the filter as a perturbation of the identity, where the second and third terms are linear and quadratic in $\textbf{D}$. For simplicity, we can elect to retain only the linear term, arriving at the approximation 
\begin{equation}
\textbf{D}^{-1} \textbf{K} \approx \textbf{I} + \textbf{D}_0^{-1}(\textbf{K}-\textbf{D}). 
\end{equation}
Letting $\textbf{D}_0 = \alpha^{-1} \textbf{I}$, we arrive at the suggested approximation in (\ref{eqn:approxfilt}). 

\textit{Choosing the best $\alpha$}: A direct approach to optimizing the value of the parameter $\alpha$ is to minimize the following cost function using the matrix {\em Frobenius} norm:
\begin{equation}
\min_{\alpha} \| \textbf{W} - \widehat{\textbf{W}}(\alpha) \|^2
\end{equation}
We can write the above difference as
\begin{equation} 
J(\alpha) = \| \textbf{W} - \widehat{\textbf{W}}(\alpha) \|^2 = \|\textbf{D}^{-1}\textbf{K} - \textbf{I}-\alpha(\textbf{K} -\textbf{D}) \|^2
\end{equation} 
This is a quadratic function in $\alpha$. Upon differentiating and setting to zero, we are led to the global minimum solution:
\begin{equation} 
\label{roots}
\widehat{\alpha} = \frac{\mbox{tr}(\textbf{K}\textbf{D}^{-1}\textbf{K}) -2\mbox{tr}(\textbf{K}) +\mbox{tr}(\textbf{D})}{\mbox{tr}(\textbf{K}^2) -2\mbox{tr}(\textbf{K} \textbf{D}) +\mbox{tr}(\textbf{D}^2)} 
\end{equation}
For sufficiently large $m$, where $m$ is the size of the widnow over which filter weights are calculated, the terms $\mbox{tr}(\textbf{D})$ and $\mbox{tr}(\textbf{D}^2)$ dominate the numerator and the denominator, respectively. Hence, 
\begin{equation}
\label{finalroot}
\widehat{\alpha} \approx \frac{\mbox{tr}(\textbf{D})}{\mbox{tr}(\textbf{D}^2)}  = \frac{s_1}{s_2},
\end{equation}
\noindent where
\begin{equation}
s_1 = \sum_{i =1}^n d_i,  \:\:\: \mbox{and} \:\:\:
s_2 = \sum_{i =1}^n d_i ^2 
\end{equation} 
This ratio is in fact bounded as $\frac{1}{mn} \leq \frac{s_1}{s_2}\leq\frac{1}{\overline{d}} $, which for large $n$ justifies a further approximation: 
\begin{equation}
\widehat{\alpha} \approx \frac{1}{\overline{d}}
\end{equation}
\noindent where $\overline{d} = \mbox{mean}(d_i)$ (the upper bound comes from the arithmetic-geometric mean inequality \cite{steelecauchy}) . Effect of this approximation on local variance is addressed in Appendix \ref{sec:appendix1}. Properties of the normalization-free filter are further discussed in \cite{normalization_free_ICIP} \footnote{Note that our proposed enhancement scheme is not dependent on the normalization-free weights, yet, using this technique can further simplify our method and result in a speed up.}. Next, our affinity-based multiscale image enhancement framework is explained.

\subsection{Proposed Filtering Scheme}
\label{sec:pfs}
\vspace{0 mm}

Our proposed filtering scheme (Fig. \ref{fig:pipeline}) has the following form:
\begin{equation}
\label{eq:proposed}
\textbf{z} = T_{1}\left(\textbf{W}_1\textbf{y}\right) + T_2\left((\textbf{W}_2 - \textbf{W}_1)\textbf{y}\right) +\cdots + T_{k}\left((\textbf{W}_{k} - \textbf{W}_{k-1})\textbf{y}\right) + T_{k+1}\left((\textbf{I}-\textbf{W}_k)\textbf{y}\right) 
\end{equation} 
where $\textbf{W}_l$ denotes the (normalized or normalization-free) filter weights with smoothing parameter $h_l$ and $T_l(.)$ is a scalar point-wise mapping function applied on each layer. It is worth mentioning that the generic filtering scheme in (\ref{eq:proposed}) includes (\ref{eqn:multi}) as a special case where $T_l(t) = \beta_l t$ and $\textbf{W}_l = \textbf{W}^{k-l+1}$. Each term in (\ref{eq:proposed}) is a filter difference applied on the input image $\textbf{y}$ and mapped through $T_l(.)$. The proposed filter consists of one high-pass term ($\textbf{I}-\textbf{W}_k$), $k-1$ band-pass terms ($\textbf{W}_{l+1} - \textbf{W}_l$) and one low-pass term ($\textbf{W}_1\textbf{y}$). Apparently the filtering behavior is determined by mapping functions $T_l(.)$ which can boost or suppress each signal layer.

\begin{figure*}[!t]
\vspace{-0 mm}
\begin{center}
\subfigure{
\includegraphics*[viewport=50 240 650 400,scale=0.65]{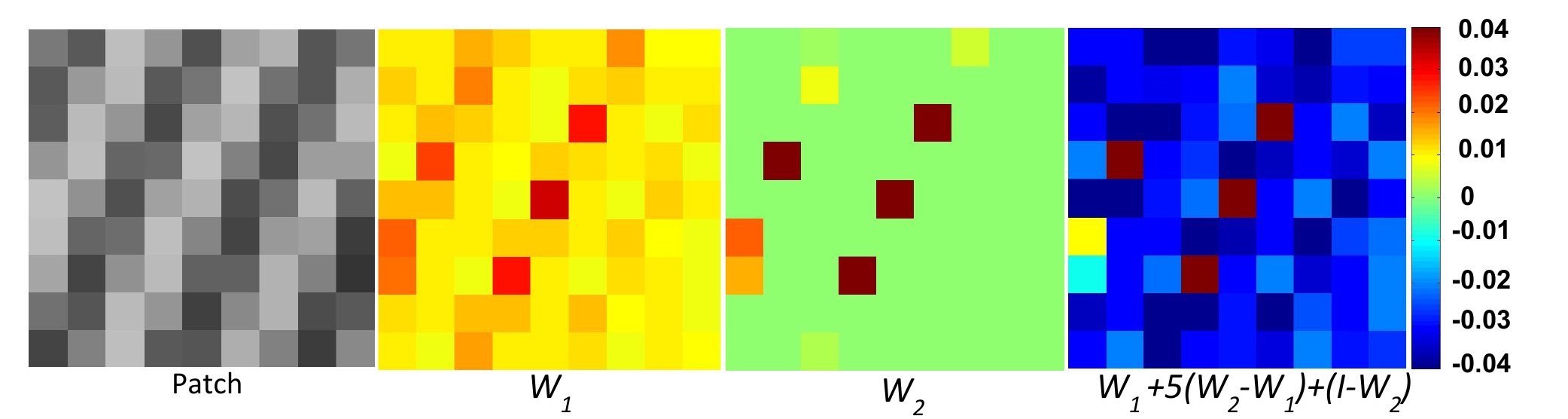}}
\end{center}
\vspace{-7 mm}
{\caption{An example of our proposed filter built on the two low-pass filters $\textbf{W}_1$ and $\textbf{W}_2$. The filter weights are shown for the center pixel of the $9\times 9$ texture patch. $\textbf{W}_1$ and $\textbf{W}_2$ contain positive affinity weights, yet the combination of their Laplacians may produce negative filter values. \label{fig:patch_filters}}}
\vspace{-0 mm}
\end{figure*}

An example of the proposed filter is shown in Fig. \ref{fig:patch_filters}. The NLM affinity weights $\textbf{W}_1$ and $\textbf{W}_2$ are computed for the center pixel in the texture patch with different smoothing parameters. The output filter is obtained by linear mapping functions as $T_1(t) = T_3(t) = t$ and $T_2(t) = 5t$. As can be seen, the output filter is a band-pass filter with both negative and positive weights.

Fig. \ref{fig:multilayer_lena} illustrates application of the proposed filtering scheme in (\ref{eq:proposed}). First, the degraded input image is decomposed into smooth and detail layers. Then, each layer is mapped by a function $T_l(.)$, and finally, all the layers are blended using a structure mask to produce the output image. As can be seen, image details are recovered in the band-pass layer and blended into the smooth layer, while the compression artifacts are suppressed. In the following, the filtering steps in (\ref{eq:proposed}) are explained in more depth.

\subsubsection{Laplacian Interpretation}
\vspace{0 mm}

Given a linear mapper as $T_l(t) = \beta_l t$, Eq. \ref{eq:proposed} can be rewritten as:
\begin{equation}
\label{eq:proposed2}
\textbf{z} = \beta_1\textbf{y} + (\beta_1-\beta_2)\textbf{L}_1\textbf{y} +\cdots +  (\beta_{k}-\beta_{k+1})\textbf{L}_k\textbf{y} 
\end{equation} 
in which $\textbf{L}_l$ represents the {\em random walk} Laplacian $\textbf{W}_l - \textbf{I}$ \cite{milanfar_SPM2011} . This basically is the input image added to a linear combination of the Laplacian-filtered images.
Another interpretation of the proposed filter can be described by {\em un-normalized} graph Laplacian \cite{milanfar_SPM2011}. As shown in Sec. \ref{sec:nff}, the normalized filter can be approximated as $\textbf{W}_l \approx \textbf{I} +  \alpha_l(\textbf{K}_l - \textbf{D}_l) = \textbf{I} + \alpha_l \mathcal{L}_l$. Then, Eq. \ref{eq:proposed2} can be expressed in terms of un-normalized Laplacians as:
\begin{equation}
\label{eq:proposed3}
\textbf{z} = \beta_1 \textbf{y} + (\beta_1-\beta_2)\alpha_1\mathcal{L}_1\textbf{y} +\cdots +  (\beta_{k}-\beta_{k+1})\alpha_k\mathcal{L}_k\textbf{y} 
\end{equation}
where $\alpha_l$ are used in the normalization approximation. Next, we address the multiple computations of the affinity weights in (\ref{eq:proposed}) .

\subsubsection{Multiple Affinity Weight Computation}
\vspace{0 mm}

The represented filtering scheme in (\ref{eq:proposed}) requires multiple computations of the edge-aware weights $\textbf{W}_l$ for $l=1,\cdots,k$. Ideally, an appropriately tuned filter based on (\ref{eq:proposed}) needs the affinity kernels to be evaluated by different smoothing parameters. This leads to a significant slow down of the algorithm's running time. This is due to the multiple evaluations of the $\exp(.)$ function. Our proposed solution to address this issue is an element-wise product of the kernel weights as:
\begin{equation}
\label{eq:weight_product}
\textbf{W}_{l+1} = \textbf{D}_{l+1}^{-1} \textbf{K}_{l+1}\:\: \left(\mbox{or}\:\: \textbf{W}_{l+1} = \alpha_{l+1} (\textbf{K}_{l+1} - \textbf{D}_{l+1})\right)\:\:\:\:\:\: \mbox{with}\:\:\:\:\:\:\: \textbf{K}_{l+1} = \textbf{K}_{l} \odot \textbf{K}_{l}
\end{equation}
where $\textbf{W}_1$ is computed explicitly, $l$ varies from $2$ to $k-1$, and $\odot$ denotes the element-wise Hadamard product. Given the exponential affinities of BL or NLM,  (\ref{eq:weight_product}) leads to a set of filters defined by smoothing parameters as $h_{l+1} =  h_{l}/2$ (both $h_x$ and $h_y$ in (\ref{eqn:bilat}) and (\ref{eqn:nlm})  will be divided by 2). In practice, we can start with a large $h_1$ and successively compute multiple versions of the filter using (\ref{eq:weight_product}).

\begin{figure*}[!t]
\vspace{-0 mm}
\begin{center}
\subfigure{
\includegraphics*[viewport=1 90 600 390,scale=0.65]{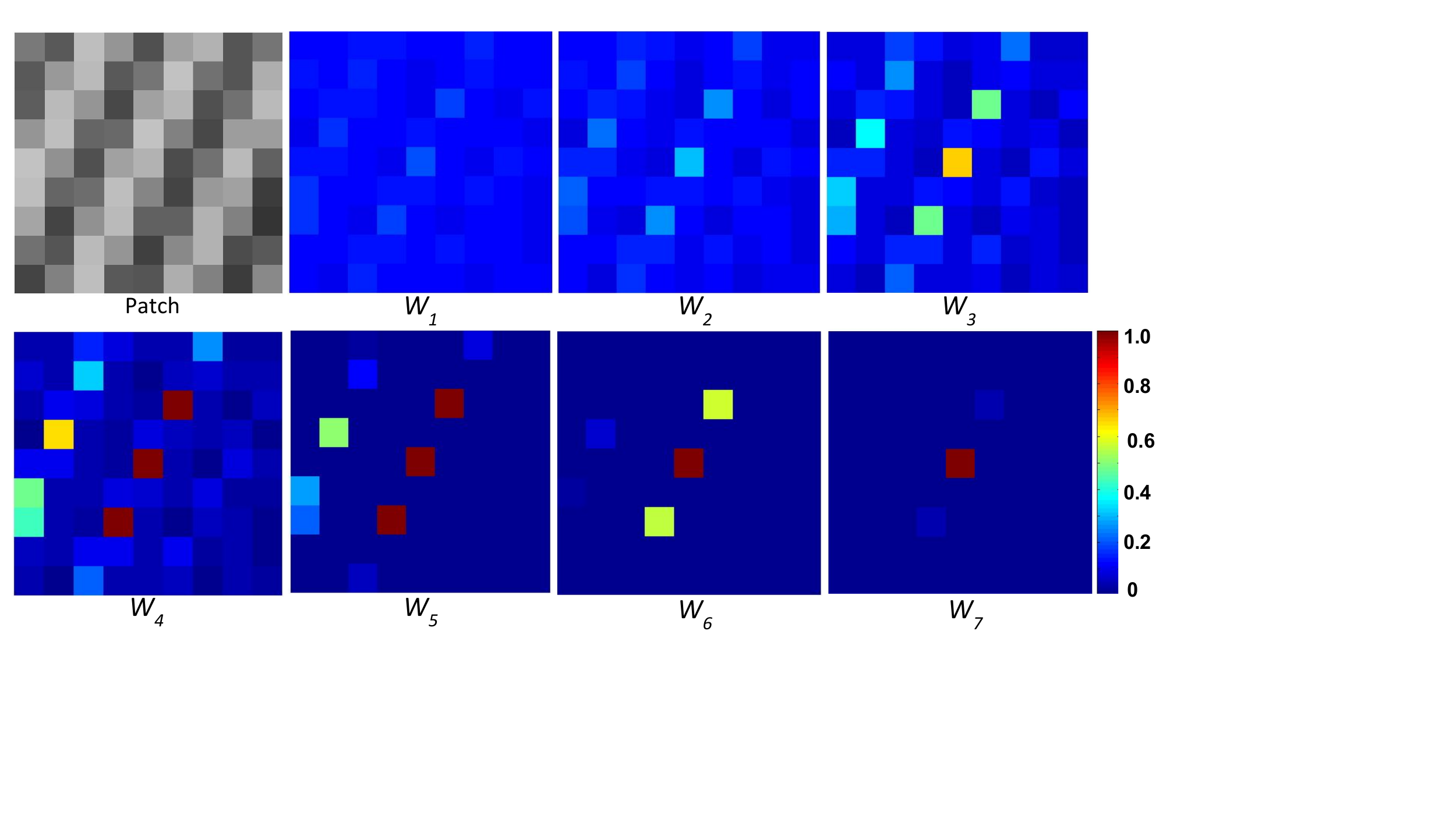}}
\end{center}
\vspace{-7 mm}
{\caption{NLM filters produced by element-wise multiplication as $\textbf{W}_{l+1} = \textbf{D}_{l+1}^{-1}\textbf{K}_{l+1}$ where $\textbf{K}_{l+1} = \textbf{K}_{l} \odot \textbf{K}_{l}$ and with $\textbf{W}_1$ computed explicitly. The filter weights are shown for the center pixel of the $9\times 9$ texture patch. \label{fig:patch_filters_multiplication}}}
\vspace{-0 mm}
\end{figure*}

An example of the element-wise weight multiplication is shown in Fig. \ref{fig:patch_filters_multiplication}. Starting with $\textbf{W}_1$, multiple filter weights from $\textbf{W}_2$ to $\textbf{W}_7$ are computed. It's worth pointing out that the variable bandwidths of these weights allow a more flexible evaluation of the proposed filtering scheme in (\ref{eq:proposed}). 

\begin{figure*}[!t]
\vspace{-0 mm}
\begin{center}
\subfigure[\scriptsize Input]{
\includegraphics*[viewport=1 1 250 250,scale=0.7]{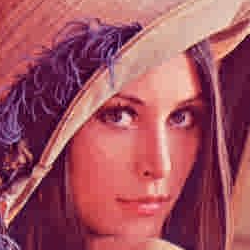}}
\subfigure[\scriptsize  $T_{1}\left(\textbf{W}_1\textbf{y}\right)$]{
\includegraphics*[viewport=1 1 250 250,scale=0.7]{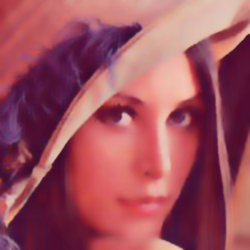}}
\subfigure[\scriptsize  $T_2\left((\textbf{W}_2 - \textbf{W}_1)\textbf{y}\right)$]{
\includegraphics*[viewport=1 1 250 250,scale=0.7]{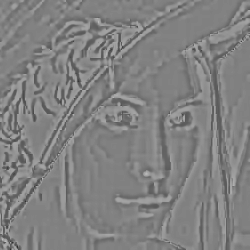}}
\subfigure[\scriptsize  $T_{1}\left(\textbf{W}_1\textbf{y}\right) + T_2\left((\textbf{W}_2 - \textbf{W}_1)\textbf{y}\right)$]{
\includegraphics*[viewport=1 1 250 250,scale=0.7]{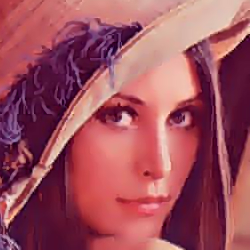}}
\end{center}
\vspace{-5 mm}
{\caption{Removing compression artifacts using our proposed method in Eq. \ref{eq:proposed}. (a) JPEG compressed image, (b) The base layer image smoothed by filter $\textbf{W}_1$, (c) The luma detail layer obtained from the band-pass filter $\textbf{W}_2 - \textbf{W}_1$, (d) Blended output $\textbf{z}$. In this example, the baseline kernel is NLM \cite{buades_nonlocal}, the mapping functions $T_l(.)$ are s-curves (see Sec. \ref{sec:mf}) and layer blending is based on a structure mask (see Sec. \ref{sec:sm}). \label{fig:multilayer_lena}}}
\vspace{-0 mm}
\end{figure*}

\subsubsection{Mapping Functions}
\label{sec:mf}
\vspace{0 mm}

\begin{figure*}[!t]
\vspace{-0 mm}
\begin{center}
\subfigure{
\includegraphics*[viewport=5 1 525 440,scale=0.28]{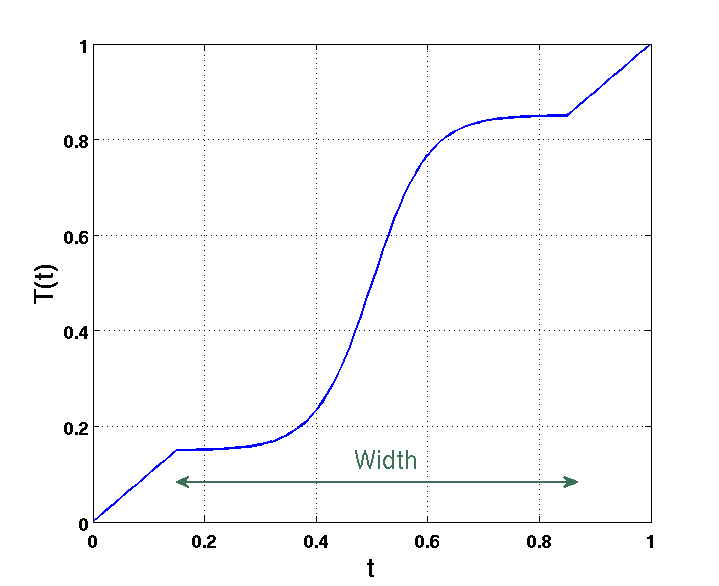}}
\subfigure{
\includegraphics*[viewport=5 1 525 440,scale=0.28]{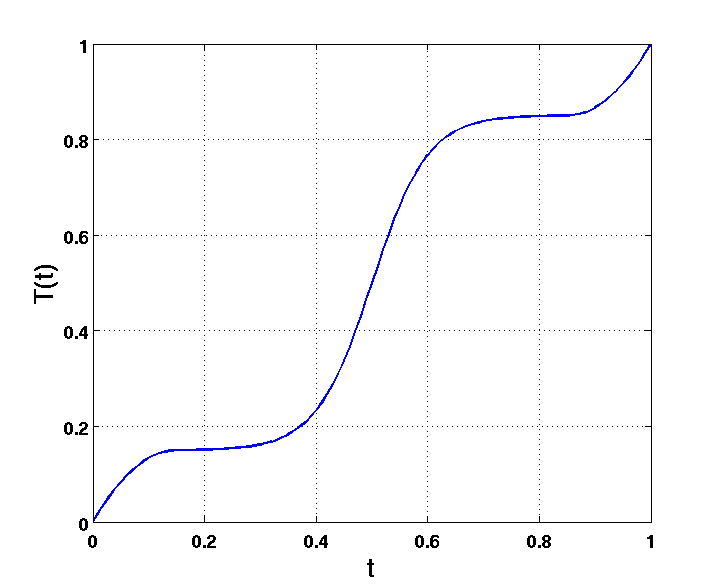} }
\subfigure{
\includegraphics*[viewport=5 1 525 440,scale=0.28]{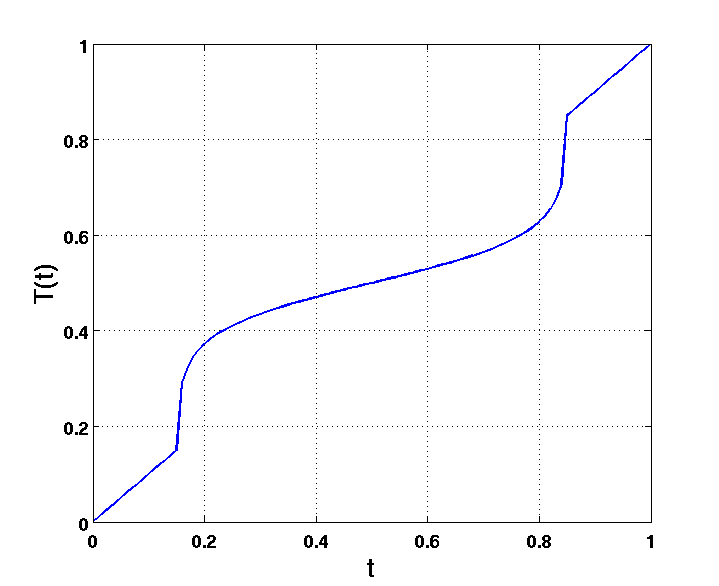} }
\end{center}
\vspace{-5 mm}
{\caption{Left: s-curve used for detail/tonal enhancement. Middle: s-curve combined with gamma correction for enhancing mid-tone contrast and boosting dark/bright details. Right: inverse s-curve used for smoothing and tone compression. \label{fig:curves}}}
\vspace{-0 mm}
\end{figure*}

The detail manipulation of the proposed algorithm strictly depends on the mapping functions $T_l(.)$. Having the input image decomposed into multiple detail layers, there are several ways to manipulate image texture and edges. The linear mapping discussed earlier is the simplest way of manipulating image details. Although the linear mapper has the interesting Laplacian interpretation, its main restrictive issue is the over-sharpening (-smoothing) of the detail content. In other words, a properly tuned detail mapping operator should treat details based on their respective local gradient magnitude. Recently, nonlinear detail manipulation has been successfully used for this task \cite{Farbman2008, Sylvain2011}. Our choice is a nonlinear mapping function, specifically the \textit{sigmoid} function:
\begin{equation}
\label{eqn:sigmoid}
T(t) = 1/(1 + \exp(-at))
\end{equation}
Our mapping operators derived from sigmoid function are demonstrated in Fig. \ref{fig:curves} (appropriate shifting and scaling is applied on the sigmoid function). Application of the s-curve mapper on the detail and base layers leads to sharpness and tonal enhancement, respectively. On the other hand, the inverse s-curve can suppress details and compress the image contrast. Given the generic sigmoid function in (\ref{eqn:sigmoid}), our mapping operator has two tuning parameters for each image layer. Parameter $a$ determines the strength of the mapping operator. The other control parameter of the mapping function is its width (illustrated in the left and right plots of Fig. \ref{fig:curves}). The width parameter can prevent generation of halo and over-sharpening artifacts around large gradient edges. It also allows mid-tone contrast enhancement without suppressing details in dark or bright regions. Another possible mapping function is the combination of gamma correction with an s-curve for enhancing dark and bright details while boosting mid-tone details (shown in the middle plot of Fig. \ref{fig:curves}). It is worth mentioning that these mapping functions can be computed in advance as look up tables and used at run time.

\begin{figure*}[!t]
\vspace{-0 mm}
\begin{center}
\subfigure[\scriptsize Input]{
\includegraphics*[scale=0.16]{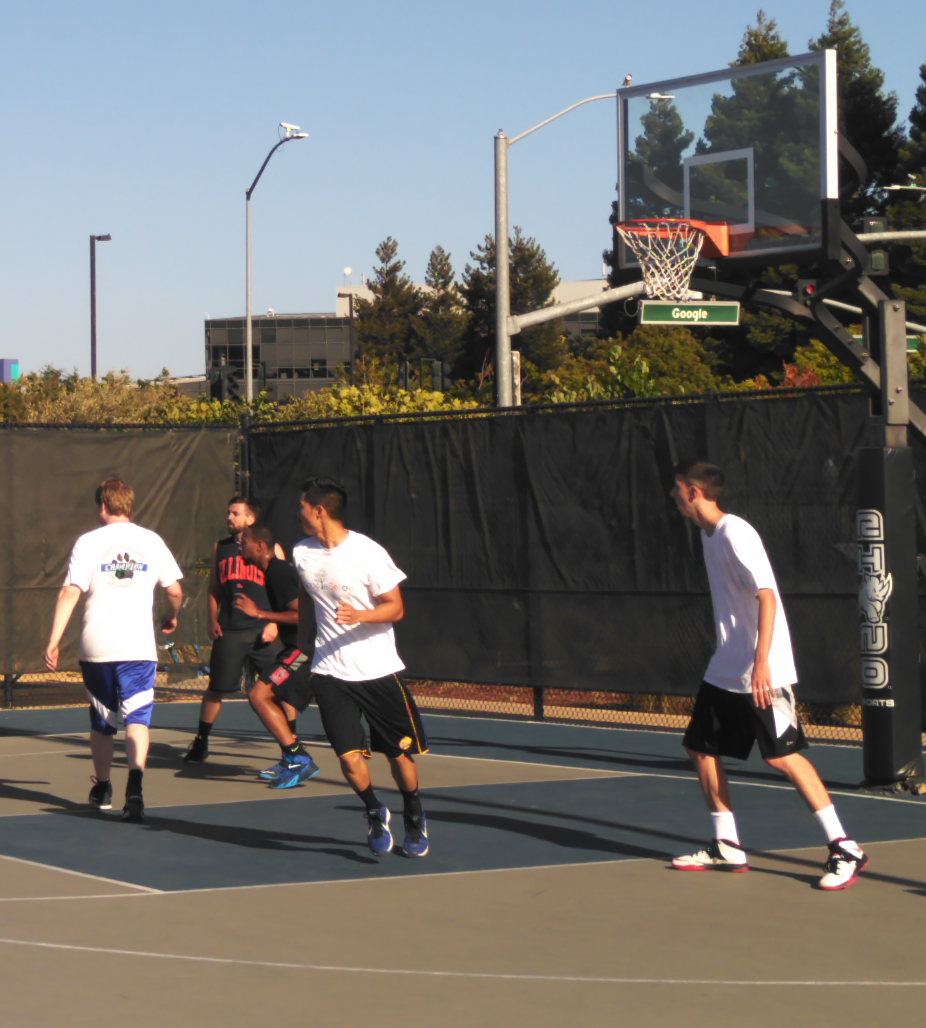}}
\subfigure[\scriptsize Enhanced (effect 1)]{
\includegraphics*[scale=0.16]{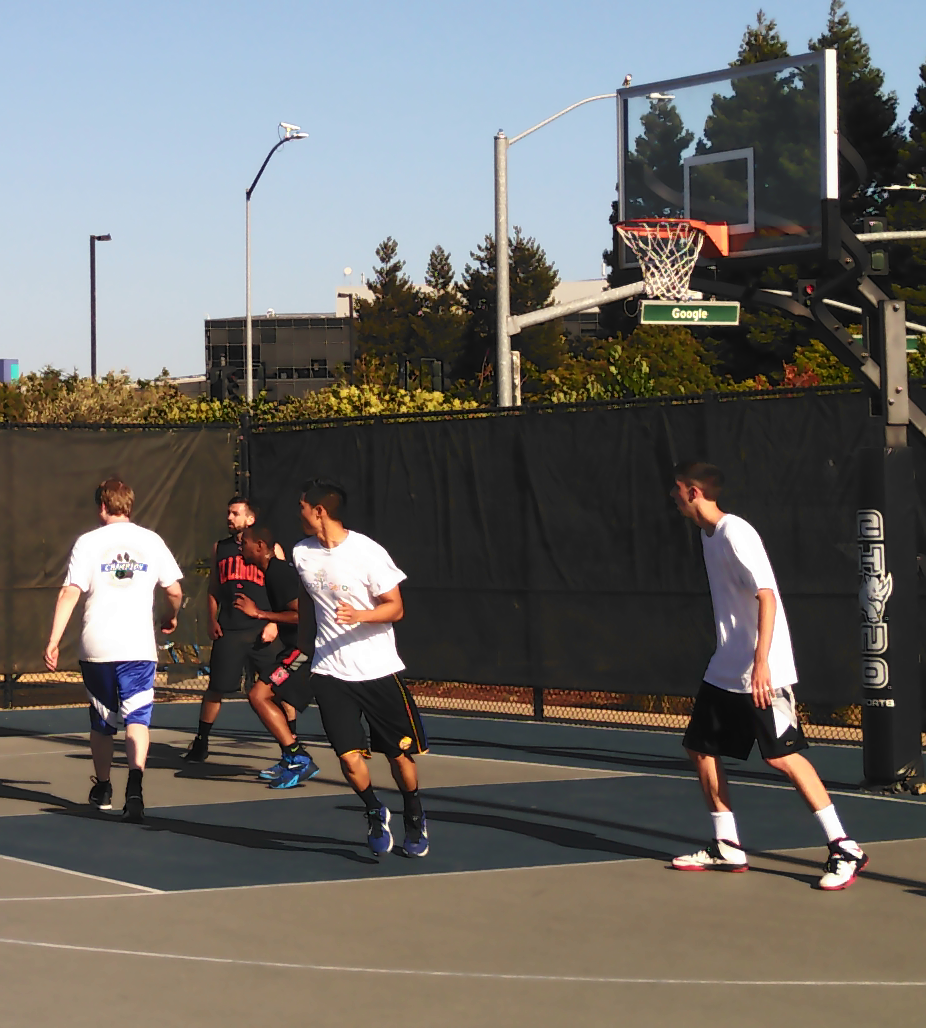}}
\subfigure[\scriptsize Enhanced (effect 2)]{
\includegraphics*[scale=0.16]{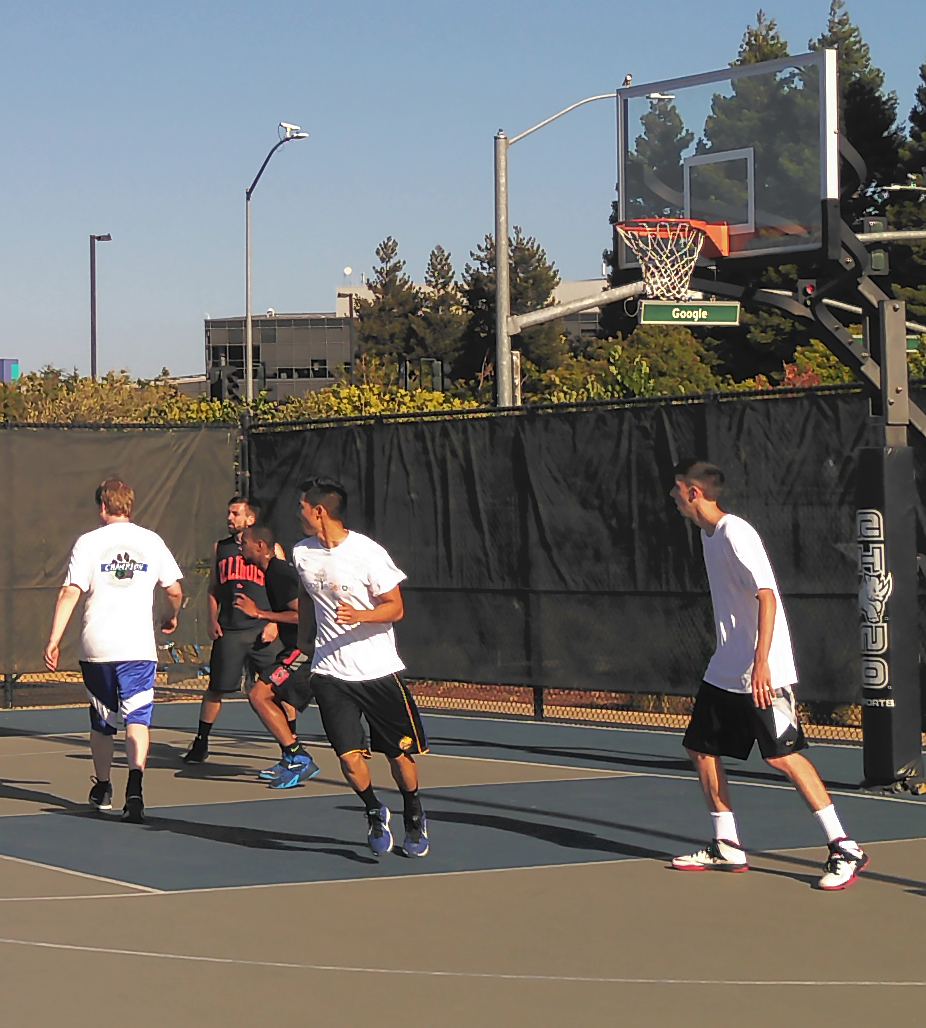}}
\end{center}
\vspace{-5 mm}
{\caption{Example of our method applied for detail enhancement. The same s-curve functions were applied on  the detail layers for both output images. The base layer image of effect 1 and effect 2 are fed to s-curve and inverse s-curve, respectively. Effect 1 represents contrast enhancement and effect 2 shows tonal compression. The input image is of size $1028\times 926$ and running time for producing the enhanced images is about $0. 031$ second. \label{fig:field}}}
\vspace{-0 mm}
\end{figure*}

Examples of applying our mapping functions are shown in Fig. \ref{fig:field}. The detail layers of both enhanced images are fed to the same s-curves, and the base (smooth) layers are fed to s-curve (effect 1) and inverse s-curves (effect 2) mappers. As can be seen, details are enhanced in both cases, with effect 1 offering higher contrast and effect 2 representing relatively lower tonal range.    

\subsubsection{Structure Mask}
\label{sec:sm}
\vspace{0 mm}

Detail enhancement and artifact magnification are inseparable. Conventionally it is preferable to boost strong image structure with high signal-to-noise (SNR) and keep the noisy regions unaltered. This requires a mechanism to detect the image structure and somehow distinguish it from other areas. Edge detection provides a rough structure mask by detecting image irregularities. However, artifacts also are prone to be recognized as image details in a gradient map. One might argue that a pre-filtered image could possibly result in a more stable edge detection; yet, this approach could lead to extra complexity in the overall framework.

\begin{figure*}[!t]
\vspace{-0 mm}
\begin{center}
\subfigure{
\includegraphics*[viewport=1130 970 1830 1800,scale=0.22]{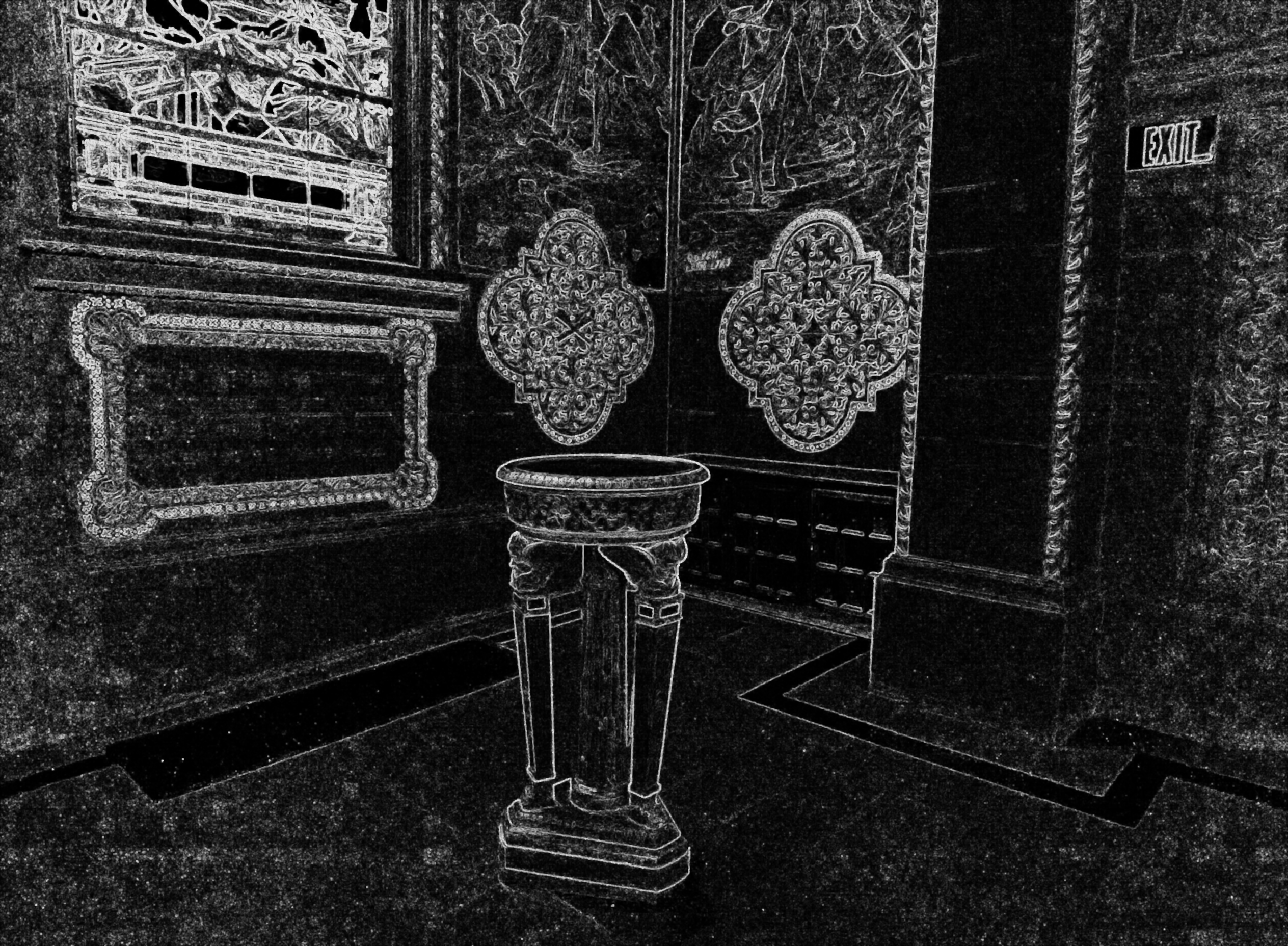}}
\subfigure{
\includegraphics*[scale=0.14]{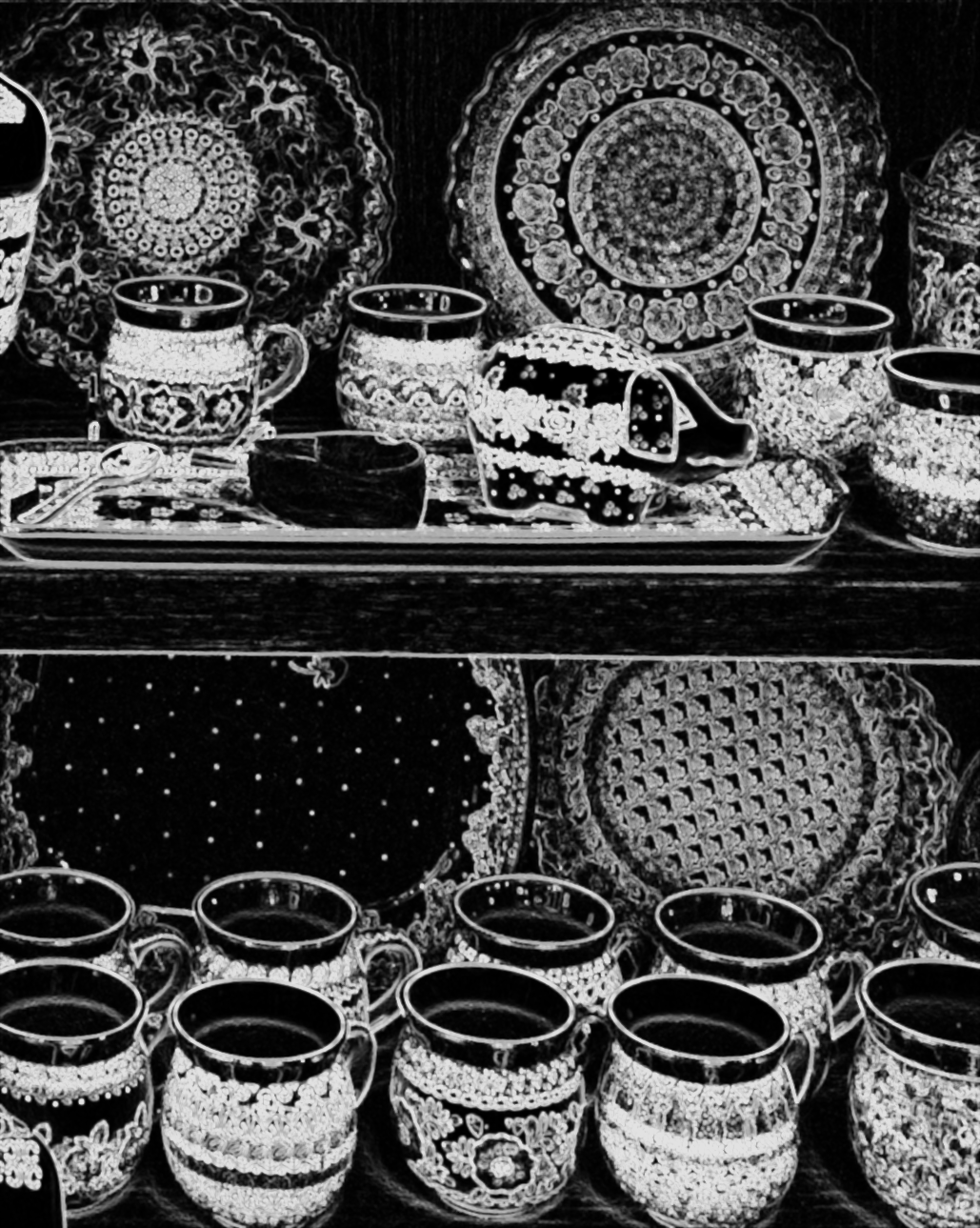}}
\end{center}
\vspace{-5 mm}
{\caption{Structure masks used for blending of the detail layers. These masks are shown for images in Fig. \ref{fig:church} and Fig. \ref{fig:dishes}.\label{fig:mask}}}
\vspace{-0 mm}
\end{figure*}

Interestingly, we have observed that the sum of the affinity degrees $[d_1, \cdots, d_p]$ (in a $p$-pixel neighborhood) conveys useful information about the image structure (see Fig.\ref{fig:OldmanDs}). A pixel located on an edge or textured region has relatively low weight sum compared to a pixel in a flat area. A soft structure mask for $i$-th pixel can be defined as:
\begin{equation}
\textbf{m}_i = 1 - d_i/p
\end{equation}
where $d_i$ denotes sum of the kernel weights associated with the $i$-th pixel and $\textbf{m}_i$ takes values in $[0, 1]$. Examples of this structure mask are demonstrated in Fig. \ref{fig:mask}. Blending results using these masks are shown in Fig. \ref{fig:church} and Fig. \ref{fig:dishes}. The detail layers of our image decomposition scheme are modulated by these masks to attenuate any possible noise and artifact boosting:
\begin{equation}
\label{eq:proposed_masked}
\textbf{z} = T_{1}\left(\textbf{W}_1\textbf{y}\right) + \textbf{M}T_2\left((\textbf{W}_2 - \textbf{W}_1)\textbf{y}\right) +\cdots + \textbf{M}T_{k}\left((\textbf{W}_{k} - \textbf{W}_{k-1})\textbf{y}\right) + \textbf{M}T_{k+1}\left((\textbf{I}-\textbf{W}_k)\textbf{y}\right) \notag
\end{equation} 
where $\textbf{M}$ is a diagonal matrix representing the structure mask with values in [0, 1]. Fig. \ref{fig:church} shows smoothing of the artifacts and sharpening of the details as a result of applying the structure mask. It is worth noting that the structure mask costs almost no additional computation, given that the kernel weights are already computed. Also, the structure mask is moderately robust to noise, because (1) it includes many summed weights, and (2) NLM kernel weights measure similarity between patches.

\section{Experimental Results}
\label{sec:results}
\vspace{0 mm}

Enhancement applications of the proposed filtering method are demonstrated through some examples in this section. NLM is our choice of affinity kernel without the spatial term given in (\ref{eqn:nlm}), and filter weights are computed in a $5\times 5$ neighborhood window. The patch size is $3\times 3$, and the smoothing parameter $h_y$ is set as $0.7$ for pixel values in $[0, 1]$. Three decomposition layers are selected based on Fig. $\ref{fig:pipeline}$ (i. e. $k=2$), meaning that NLM weights are computed once ($\textbf{W}_1$), and used in the element-wise weight multiplication to form the second filter ($\textbf{W}_2$).

\begin{figure*}[!t]
\vspace{-0 mm}
\begin{center}
\subfigure[\scriptsize Input]{
\includegraphics*[viewport=100 350 420 700,scale=0.45]{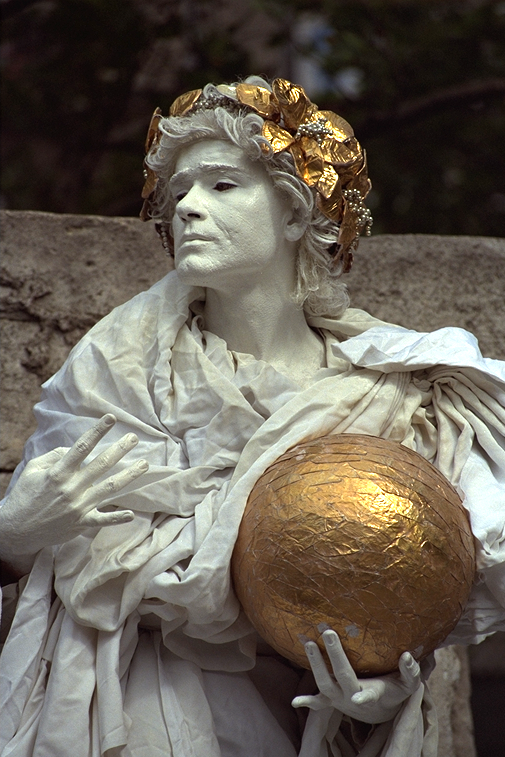}}
\subfigure[\scriptsize  \cite{GIF}]{
\includegraphics*[viewport=100 350 420 700,scale=0.45]{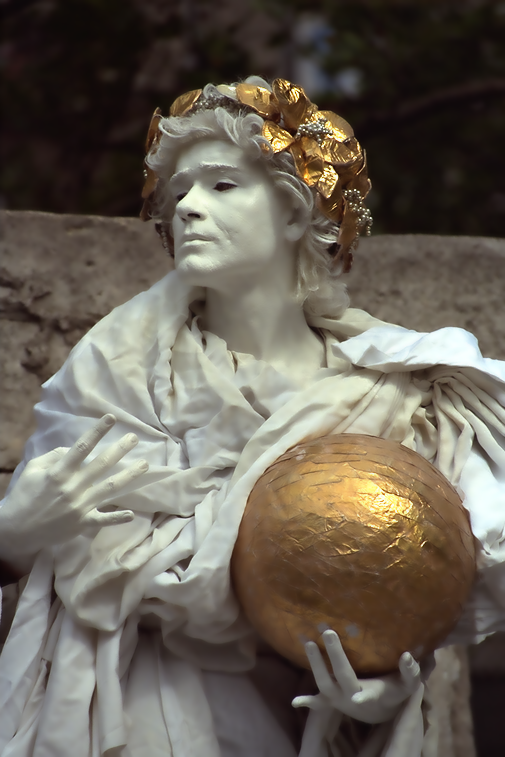}}
\subfigure[\scriptsize  Ours]{
\includegraphics*[viewport=100 350 420 700,scale=0.45]{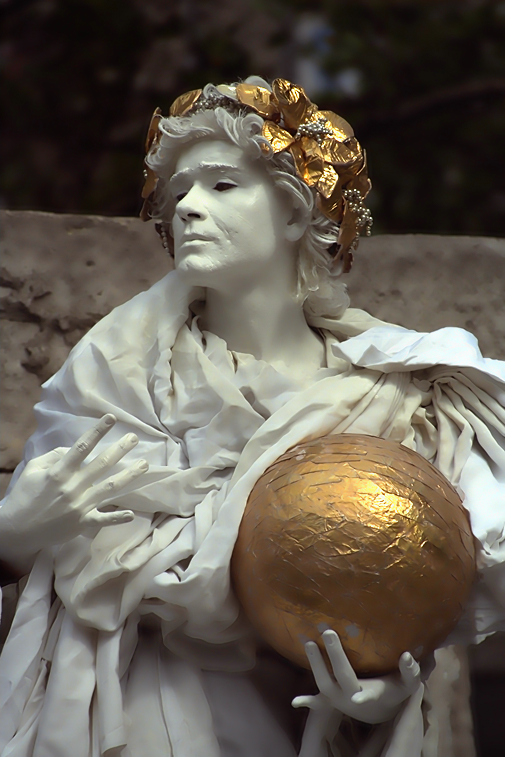}}
\end{center}
\vspace{-5 mm}
{\caption{Edge-aware smoothing using our method compared to the result from guided image filtering \cite{GIF}. For nearly the same running time budget (0.08 second), our method better flattens the fine details and avoids blurring the piecewise smooth output. \label{fig:smoothing1}}}
\vspace{-0 mm}
\end{figure*}

\begin{figure*}[!t]
\vspace{-0 mm}
\begin{center}
\subfigure[\scriptsize Input]{
\includegraphics*[viewport=120 100 480 520,scale=0.4]{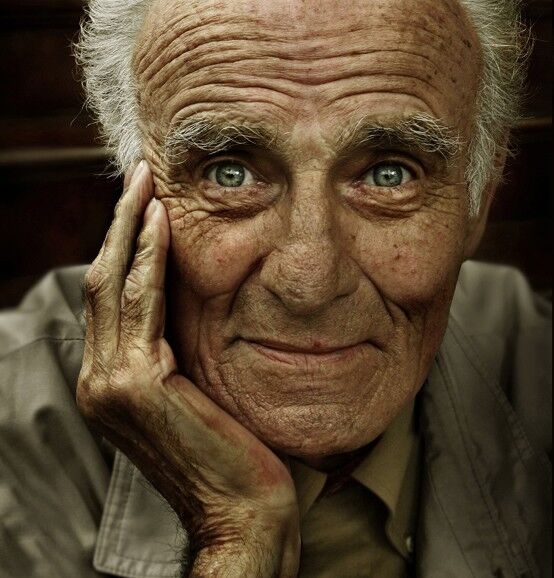}}
\subfigure[\scriptsize  \cite{GIF}]{
\includegraphics*[viewport=120 100 480 520,scale=0.4]{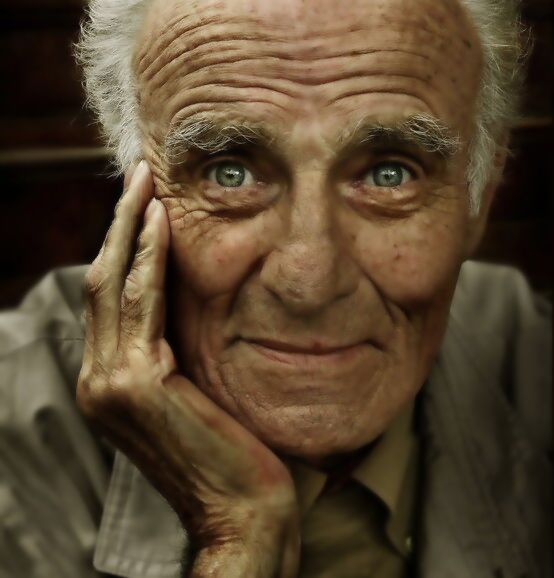}}
\subfigure[\scriptsize  Ours]{
\includegraphics*[viewport=120 100 480 520,scale=0.4]{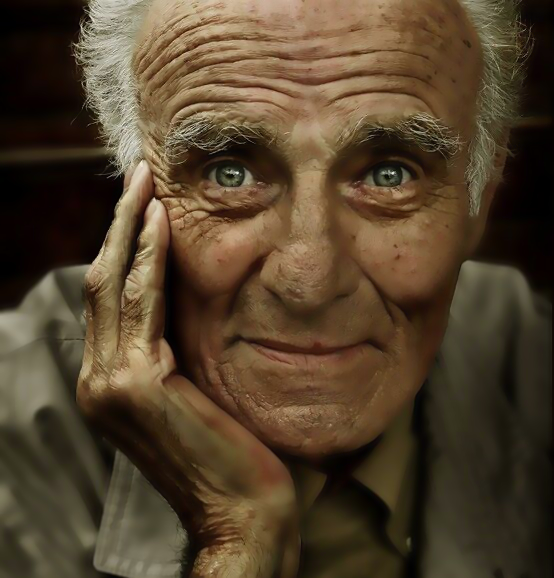}}
\end{center}
\vspace{-5 mm}
{\caption{Edge-aware smoothing using our method compared to the result from guided image filtering \cite{GIF}. For nearly the same running time budget (0.06 second), our method better flattens the fine details and avoids blurring the piecewise smooth output. \label{fig:smoothing2}}}
\vspace{-0 mm}
\end{figure*}

\begin{figure*}[!t]
\vspace{-0 mm}
\begin{center}
\subfigure[\scriptsize Input]{
\includegraphics*[scale=0.22]{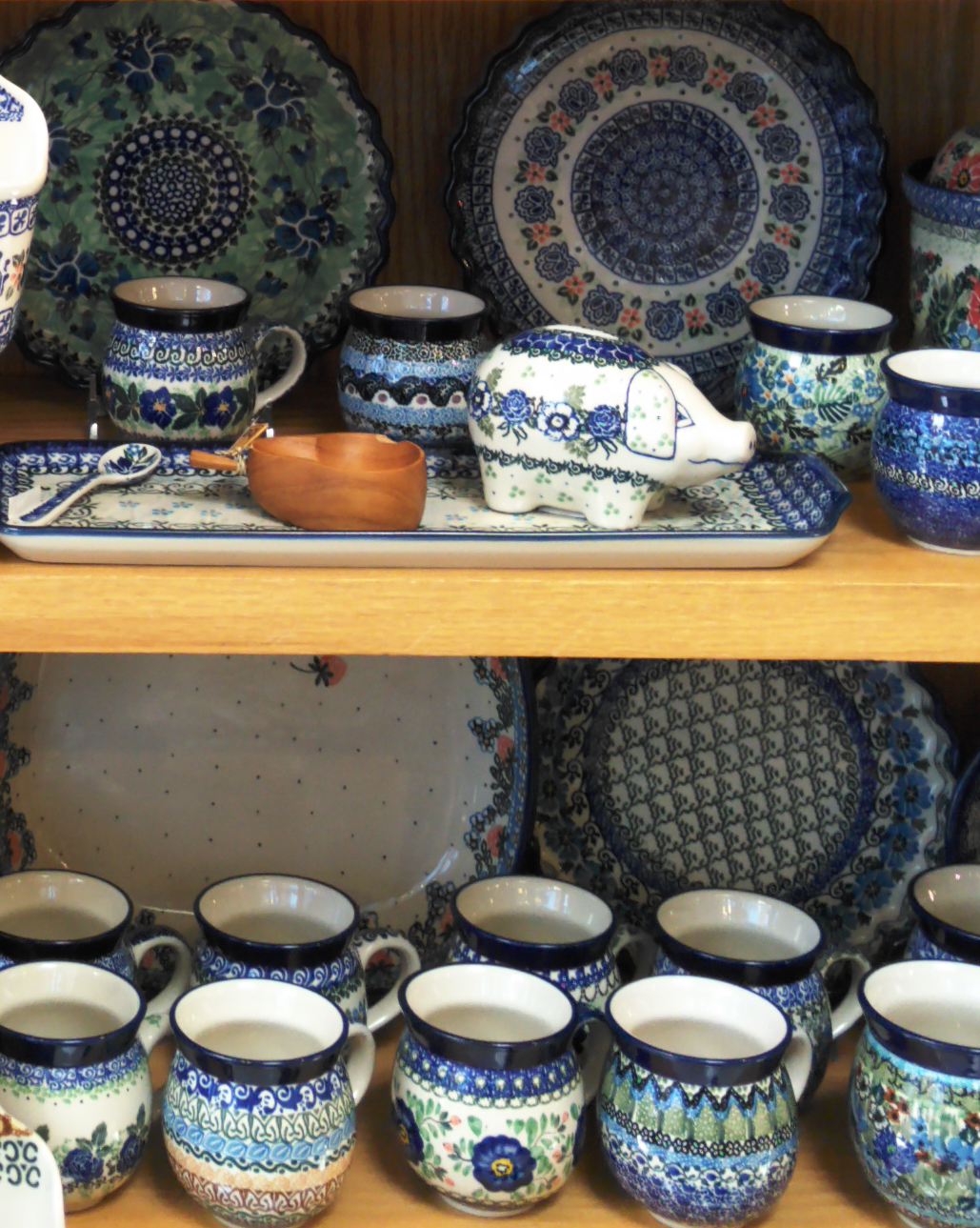}}
\subfigure[\scriptsize Enhanced details]{
\includegraphics*[scale=0.22]{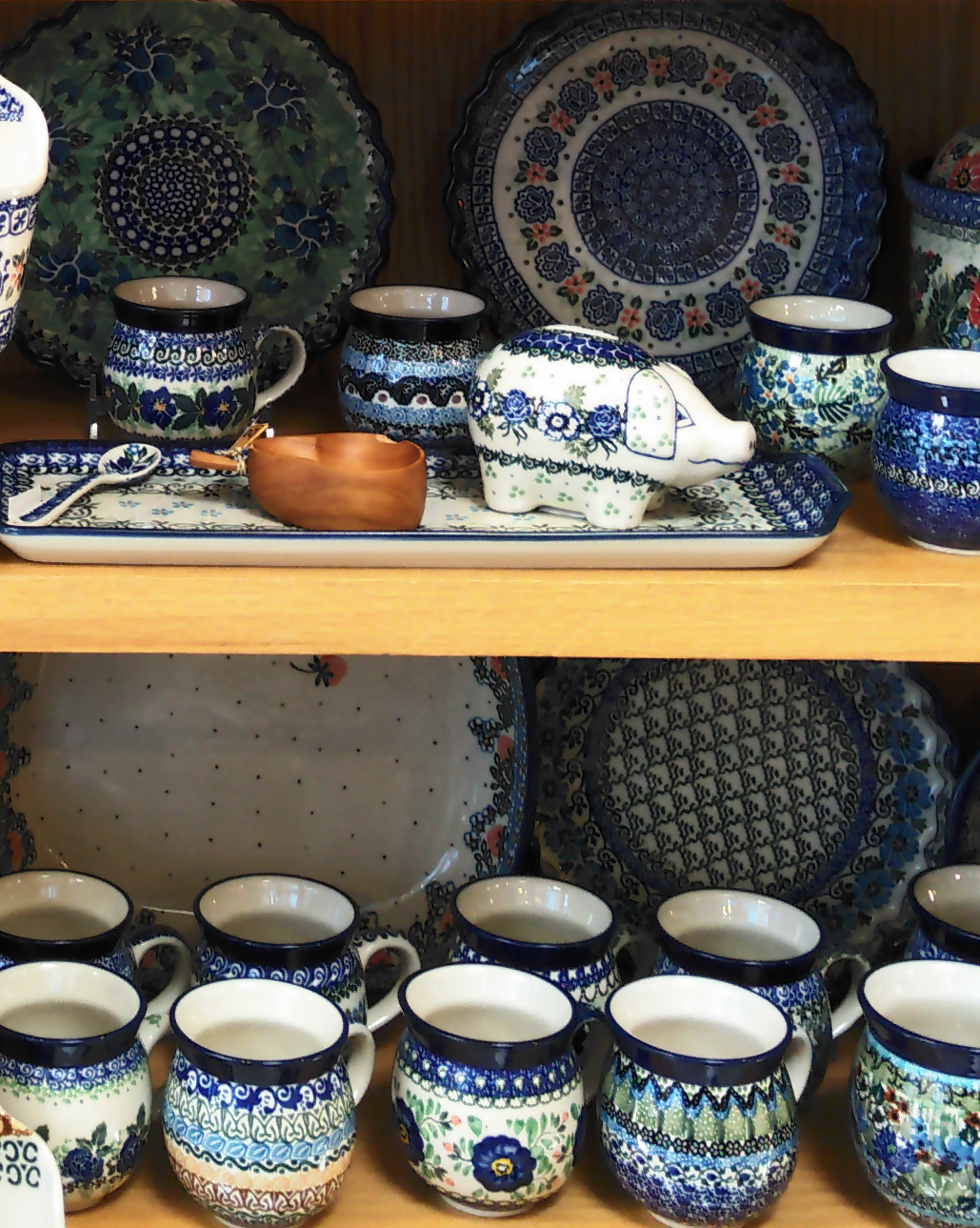} }
\end{center}
\vspace{-5 mm}
{\caption{Example of our method applied for detail enhancement. The input image is of size $1289\times 1029$ and running time for producing the enhanced image is about $0. 04$ second. \label{fig:dishes}}}
\vspace{-0 mm}
\end{figure*}

\begin{figure*}[!t]
\vspace{-0 mm}
\begin{center}
\subfigure[\scriptsize Input]{
\includegraphics*[viewport=1 1 200 550,scale=0.43]{./figures/dishes.png}}
\subfigure[\scriptsize  \cite{AUM}]{
\includegraphics*[viewport=1 1 200 550,scale=0.43]{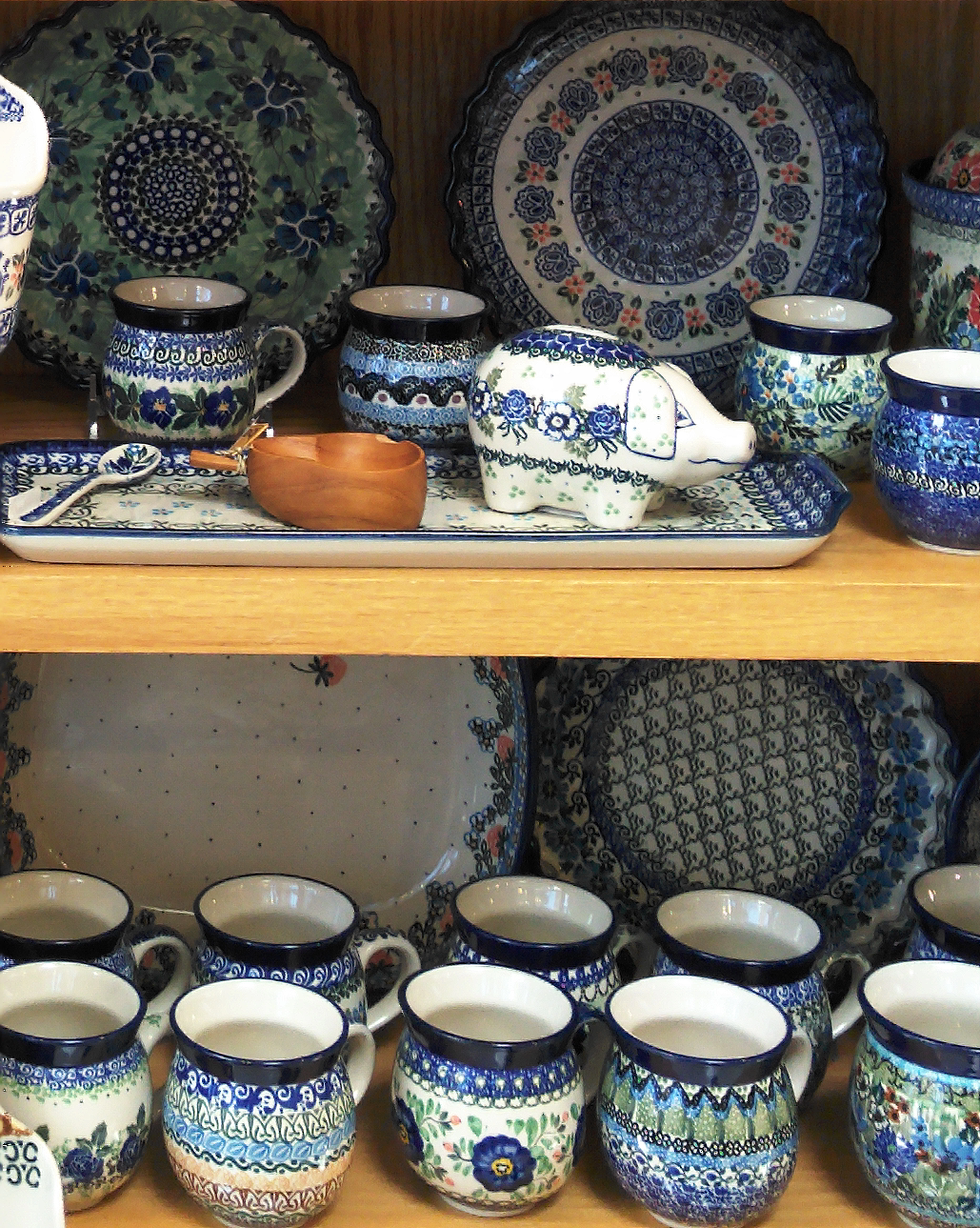}}
\subfigure[\scriptsize  \cite{C_UM}]{
\includegraphics*[viewport=1 1 200 550,scale=0.43]{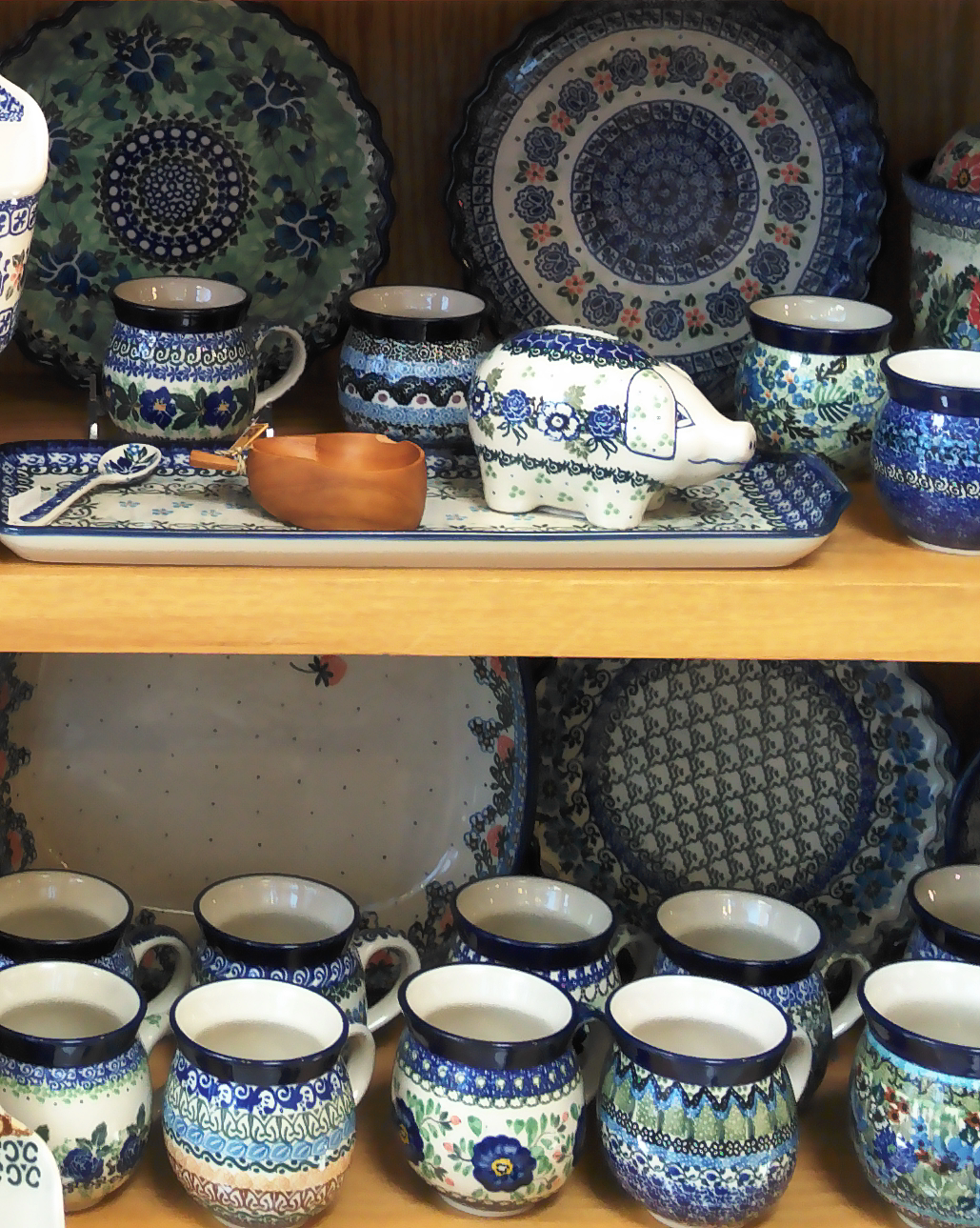}}
\subfigure[\scriptsize  \cite{Farbman2008}]{
\includegraphics*[viewport=1 1 200 550,scale=0.43]{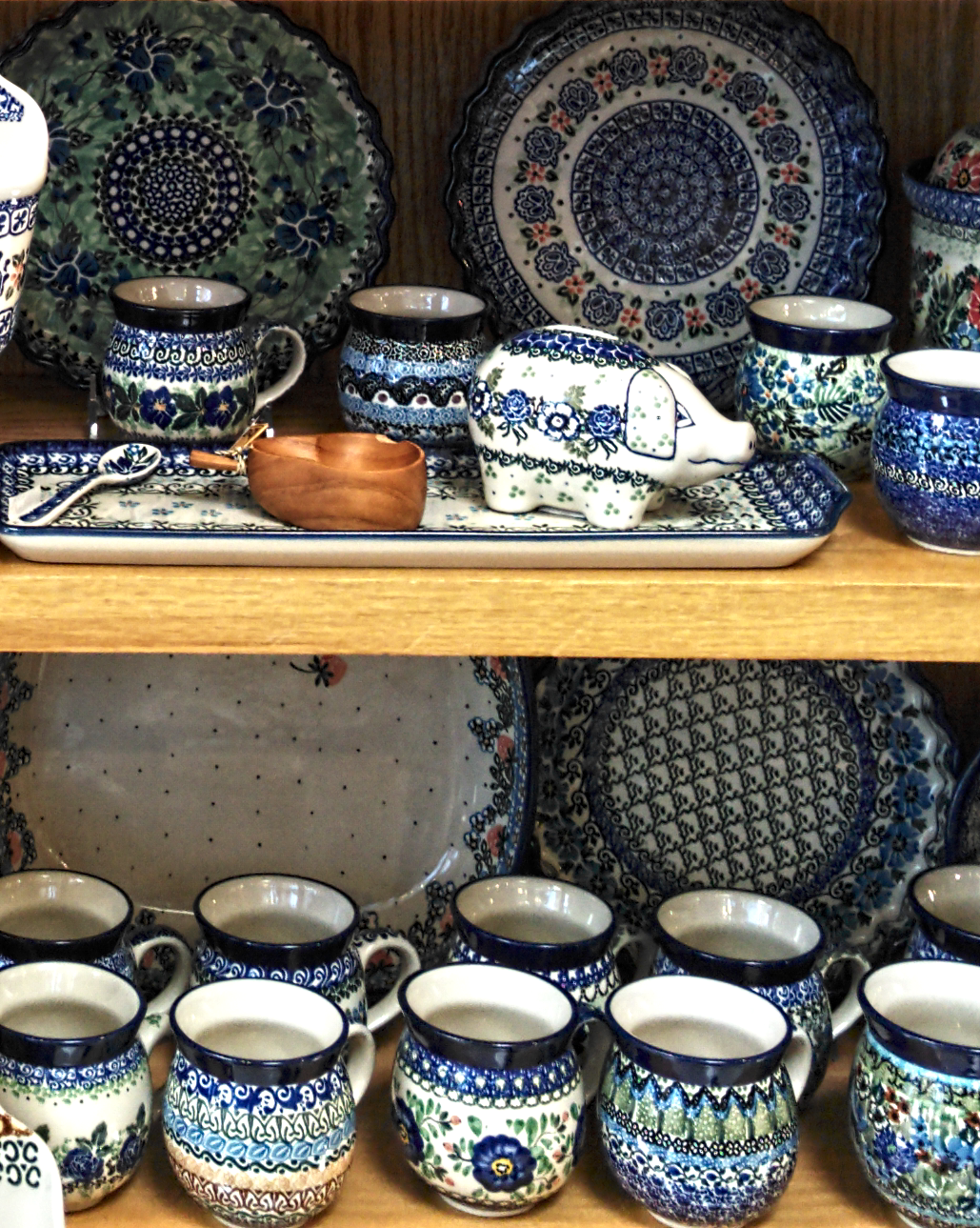}}
\subfigure[\scriptsize Ours]{
\includegraphics*[viewport=1 1 200 550,scale=0.43]{./figures/dishes_enhance.png}}
\end{center}
\vspace{-5 mm}
{\caption{Comparing existing detail enhancement methods with our proposed algorithm. \label{fig:comparison1}}}
\vspace{-0 mm}
\end{figure*}

\begin{figure*}[!t]
\vspace{-0 mm}
\begin{center}
\subfigure[\scriptsize Input]{
\includegraphics*[scale=0.75]{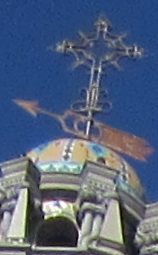}}
\subfigure[\scriptsize  \cite{AUM}]{
\includegraphics*[scale=0.75]{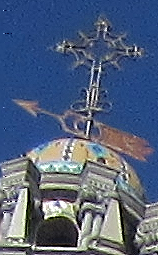}}
\subfigure[\scriptsize  \cite{C_UM}]{
\includegraphics*[scale=0.75]{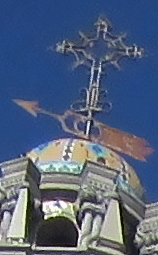}}
\subfigure[\scriptsize  \cite{multi_retinex}]{
\includegraphics*[scale=0.75]{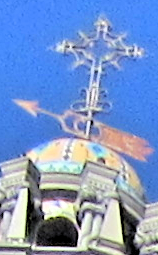}}
\subfigure[\scriptsize  \cite{Farbman2008}]{
\includegraphics*[scale=0.75]{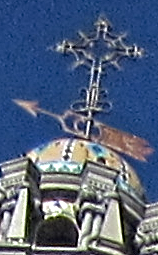}}
\subfigure[\scriptsize Ours]{
\includegraphics*[scale=0.75]{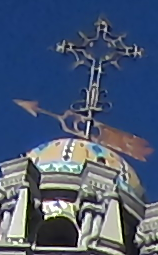}}
\end{center}
\vspace{-5 mm}
{\caption{Comparing existing detail enhancement methods with our proposed algorithm. Running times (in seconds): (b) 1.86, (c) 0.69, (d) 0.05, (e) 0.19, (f) 0.002.\label{fig:comparison2}}}
\vspace{-0 mm}
\end{figure*}

\begin{figure*}[!t]
\vspace{-0 mm}
\begin{center}
\subfigure[\scriptsize Input]{
\includegraphics*[viewport=60 1 233 244,scale=0.7]{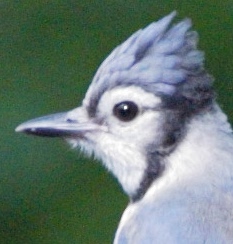}}
\subfigure[\scriptsize  \cite{AUM}]{
\includegraphics*[viewport=60 1 233 244,scale=0.7]{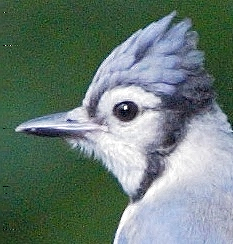}}
\subfigure[\scriptsize  \cite{C_UM}]{
\includegraphics*[viewport=60 1 233 244,scale=0.7]{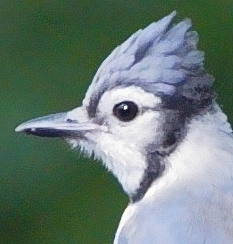}}
\subfigure[\scriptsize  \cite{multi_retinex}]{
\includegraphics*[viewport=60 1 233 244,scale=0.7]{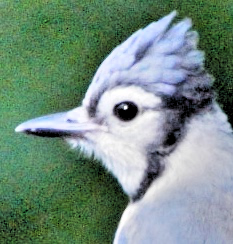}}
\subfigure[\scriptsize  \cite{Farbman2008}]{
\includegraphics*[viewport=60 1 233 244,scale=0.7]{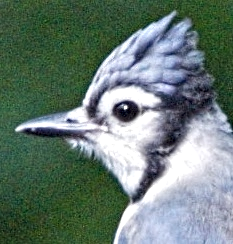}}
\subfigure[\scriptsize Ours]{
\includegraphics*[viewport=60 1 233 244,scale=0.7]{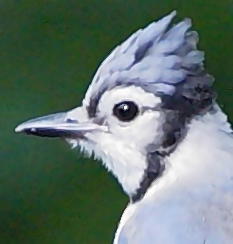}}
\end{center}
\vspace{-4 mm}
{\caption{Comparing existing detail enhancement methods with our proposed algorithm. Running times (in seconds): (b) 2.83, (c) 1.16, (d) 0.08, (e) 0.031, (f) 0.003. \label{fig:comparison3}}}
\vspace{-0 mm}
\end{figure*}

\begin{figure*}[!t]
\vspace{-0 mm}
\begin{center}
\subfigure[\scriptsize Input]{
\includegraphics*[viewport=450 200 700 800,scale=0.34]{./figures/church.png}}
\subfigure[\scriptsize  \cite{AUM}]{
\includegraphics*[viewport=450 200 700 800,scale=0.34]{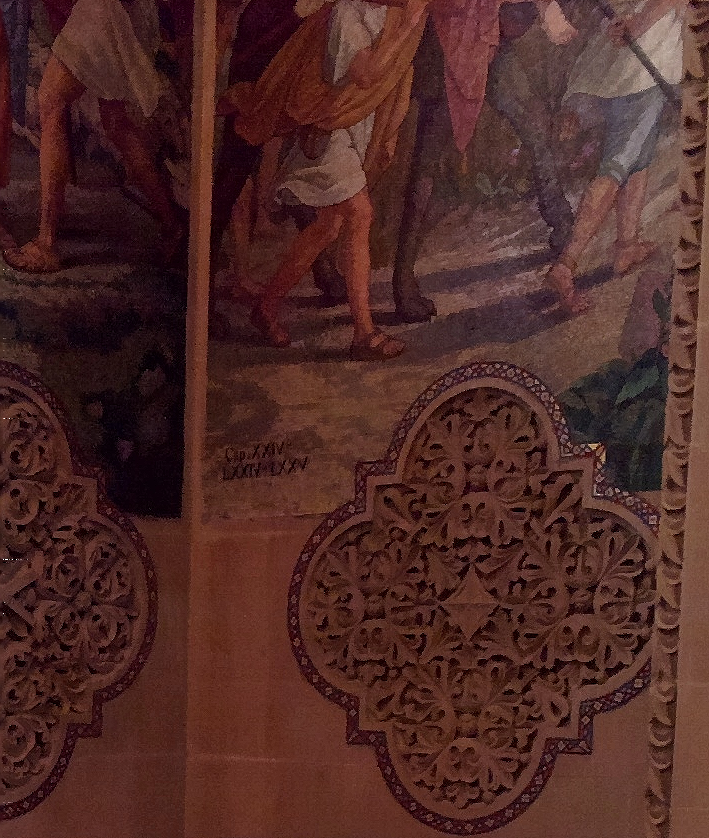}}
\subfigure[\scriptsize  \cite{C_UM}]{
\includegraphics*[viewport=450 200 700 800,scale=0.34]{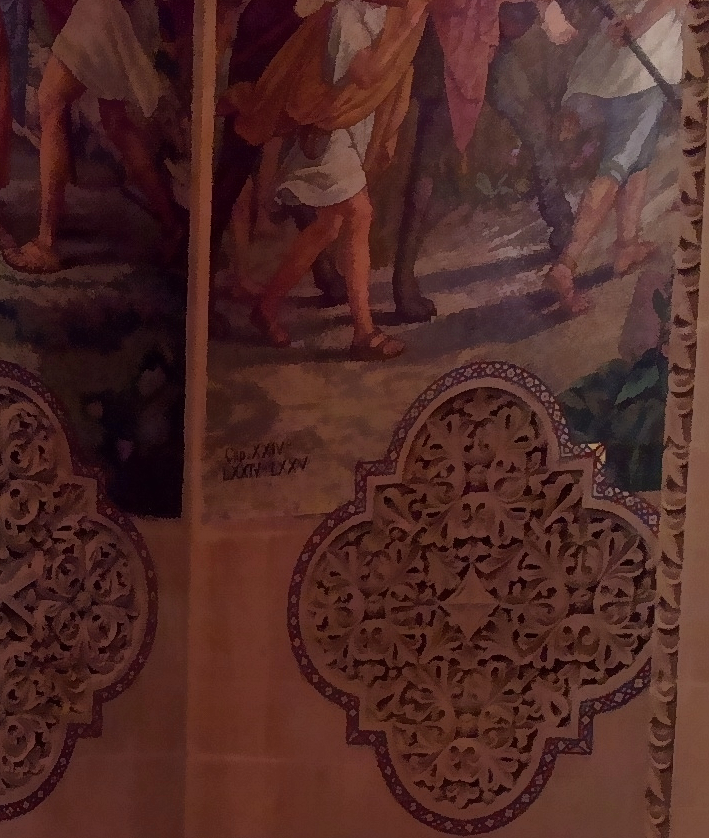}}
\subfigure[\scriptsize  \cite{Farbman2008}]{
\includegraphics*[viewport=450 200 700 800,scale=0.34]{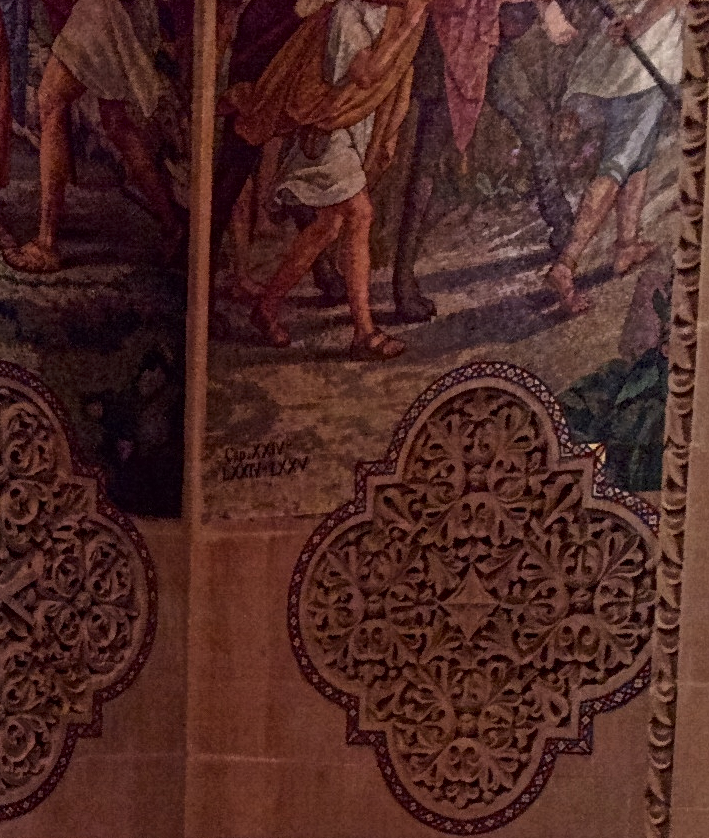}}
\subfigure[\scriptsize Ours]{
\includegraphics*[viewport=450 200 700 800,scale=0.34]{./figures/church_enhance.png}}
\end{center}
\vspace{-4 mm}
{\caption{Comparing existing detail enhancement methods with our proposed algorithm. \label{fig:comparison4}}}
\vspace{-0 mm}
\end{figure*}

There are three main applications for our method. First, detail smoothing (shown in Fig. \ref{fig:smoothing1} and Fig. \ref{fig:smoothing2}); second, sharpening mildly blurred images (shown in Fig. \ref{fig:dishes}, and Fig. \ref{fig:comparison1}), and finally, detail enhancement in noisy/artifacted images (shown in Fig. \ref{fig:comparison2}, Fig. \ref{fig:comparison3} and Fig. \ref{fig:comparison4}). Mapping functions $T_l(.)$ are tuned specifically for each application to produce the best results. 

Our multiscale decomposition allows smoothening fine details while preserving medium and coarse scale details. Our method is compared to the guided edge-aware filter \cite{GIF} in Fig. \ref{fig:smoothing1} and Fig. \ref{fig:smoothing2}. These results are obtained by removing the fine scale detail layer and mapping the medium scale layer (see Fig. \ref{fig:pipeline}) by an s-curve of width $0.2$ and $a=10$. As can be seen, in contrast to the guided filter, our result is less blurry.

Out-of-focus blur is another common problem in mobile imaging. Objects typically lose sharpness and local contrast in a mildly blurred scene (see input photo in Fig. \ref{fig:dishes}). Our filtering framework can effectively enhance these images (see output photo in Fig. \ref{fig:dishes}). Parameters of the s-curve functions in each scale are: $a=20$ and width of $0.66$ for the fine scale detail layer, $a=50$ and width of $0.33$ for the medium scale detail layer, and $a=6$ with width of $0.75$ for the base layer. Comparison of the proposed method with other techniques is demonstrated in Fig. \ref{fig:comparison1}. The adaptive unsharp masking \cite{AUM} and Farbman's detail enhancement \cite{Farbman2008} tend to boost the image sharpness and noise together. Our result demonstrates better local contrast with no noise magnification or detail loss.

\begin{figure*}[!t]
\vspace{-0 mm}
\begin{center}
\subfigure[\scriptsize Input]{
\includegraphics*[viewport=80 160 300 470,scale=0.49]{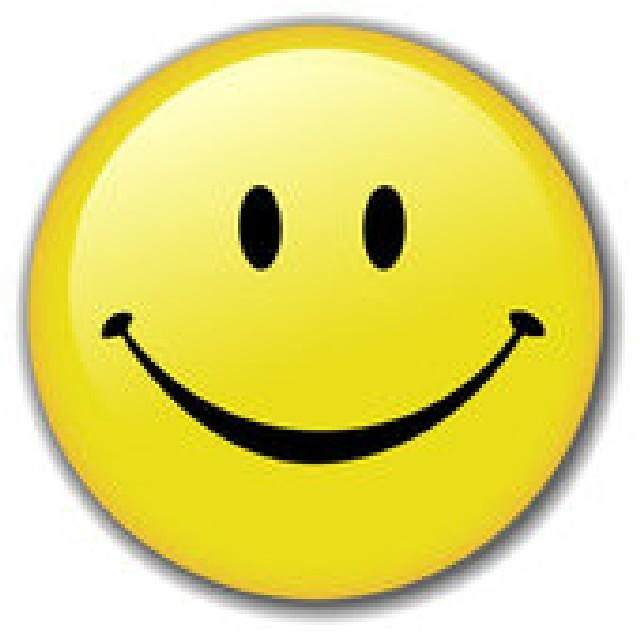}}
\subfigure[\scriptsize  Ours]{
\includegraphics*[viewport=80 160 300 470,scale=0.49]{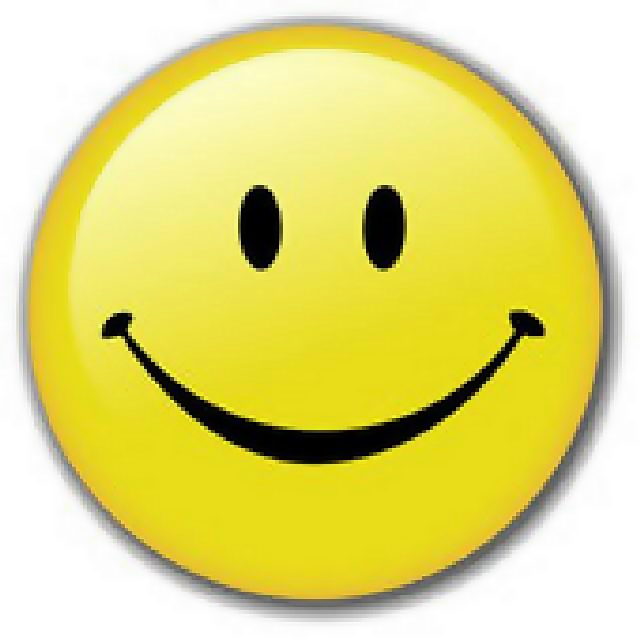}}
\subfigure[\scriptsize Input]{
\includegraphics*[viewport=1 50 150 260,scale=0.72]{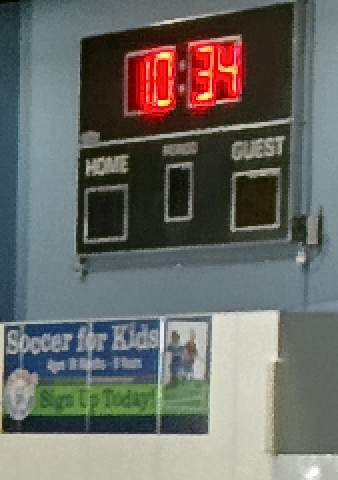}}
\subfigure[\scriptsize  Ours]{
\includegraphics*[viewport=1 50 150 260,scale=0.72]{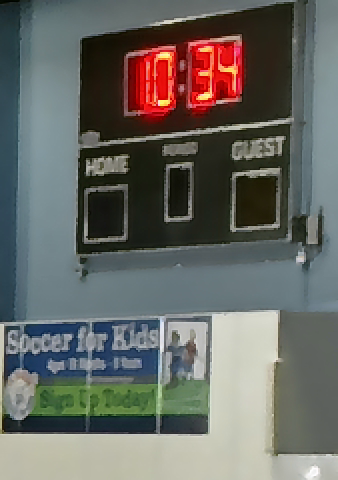}}
\end{center}
\vspace{-4 mm}
{\caption{Removing compression artifacts using our method. Filters are applied in RGB domain and are computed in an $11\times 11$ neighborhood window. \label{fig:comparison6}}}
\vspace{-0 mm}
\end{figure*}

\begin{figure*}[!t]
\vspace{-0 mm}
\begin{center}
\subfigure[\scriptsize Input]{
\includegraphics*[viewport=5 5 105 80,scale=0.95]{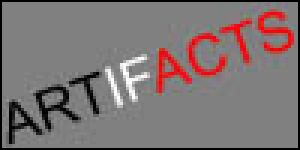}}
\subfigure[\scriptsize \cite{GIF}]{
\includegraphics*[viewport=5 5 105 80,scale=0.95]{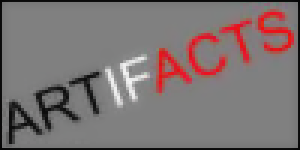}}
\subfigure[\scriptsize  Ours]{
\includegraphics*[viewport=5 5 105 80,scale=0.95]{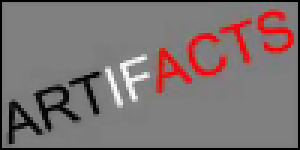}}
\end{center}
\vspace{-4 mm}
{\caption{Removing compression artifacts using (b) guided filter \cite{GIF} and (c) our method. Both filters are applied in RGB domain and are computed in an $11\times 11$ neighborhood window. For the same running time budget, our result is superior to \cite{GIF}. \label{fig:comparison5}}}
\vspace{-0 mm}
\end{figure*}

Noise is an inevitable part of any imaging pipeline. We also used our method for enhancing images corrupted by real noise and other artifacts (see input images in Fig. \ref{fig:multilayer_lena}, Fig. \ref{fig:comparison2}, Fig. \ref{fig:comparison3}, Fig. \ref{fig:comparison4}, Fig. \ref{fig:comparison5} and Fig. \ref{fig:comparison6}). To better handle noise in the input image, the fine scale detail is suppressed in our image decomposition and the base and medium scale layers are boosted. The mapping parameters to achieve this effect are: $a=10$ and width of $1$ for the fine scale detail layer (inverse s-curve), $a=60$ and width of $0.45$ for the medium scale detail layer (s-curve), and $a=5$ and width of $0.75$ for the base layer (s-curve). Fig. \ref{fig:comparison2}-\ref{fig:comparison6} show examples of noisy/artifacted images enhanced by different methods. Overall, visual comparisons indicate superiority of the proposed algorithm when dealing with degraded images.

Our C++ implementation is tested on an Intel Xeon CPU @ 3.5 GHz with 32 MB memory. Complexity of the proposed algorithm is linearly dependent on the filter size. Running time of our method is reported for some test images in Table \ref{tab:times}. Examples shown in this paper are mostly based on $5\times 5$ NLM filters, leading to an average speed of 21 MP/sec. Given available implementations of the other enhancement techniques, our method is significantly faster. For instance, processing an image of size 0.5 Mega pixel takes 0.03, 0.91, 3.2, 30.5, 12.7 seconds for \cite{GIF}, \cite{multi_retinex}, \cite{Farbman2008}, \cite{AUM}, and \cite{C_UM} , respectively. Our implementation takes less than 0.025 seconds to enhance the same image. We also tested our algorithm without weight normalization and weight re-computation approximations to measure the overall saved time. Our experiments suggest that these approximations lower the running time by 15-20\%.

\begin{table*}[!t]
\vspace{0 mm}
\begin{center}
\caption{Average running time (seconds) of the proposed algorithm computed for NLM kernel of different sizes. Size of the test images along with the neighborhood window sizes are shown in the first row and column of the table.}
\scalebox{0.95}{\begin{tabular}[c]{|c||c|c|c|c|c|c|}
\hline
 {} &{\bf 0.4 MP} & {\bf 1 MP} &{\bf 3 MP} & {\bf 12 MP} \\
\hline
$3\times3$ & 0.014 & 0.019 &  0.034 & 0.143\\
\hline
$5\times5$ & 0.022 & 0.045 &  0.105 & 0.575\\
\hline
$7\times7$ & 0.040 & 0.075 &  0.223 & 1.363\\
\hline
$9\times9$ & 0.078 & 0.152 &  0.473 & 2.623\\
\hline
\end{tabular}} \\
\label{tab:times}
\end{center}
\vspace{-2 mm}
\end{table*}

\section{Conclusion}
\label{sec:conclusion}
\vspace{0 mm}

We introduced a new multiscale image enhancement algorithm to improve on the existing edge-aware filters. Our multiscale decomposition scheme provides a fast detail manipulation paradigm with a minor complexity added to the computation of the baseline kernel. Combination of the detail layers with a structure mask produces state-of-the-art image enhancement results, addressing shortcomings of the existing algorithms. This proposed work is implemented for NLM filter weights; however, it can be easily extended to other edge-aware filters. 

\appendices
\section{Effect of Approximation on Local Variance}
\label{sec:appendix1}

We expect that the approximate filter should affect the variance of the output pixels. Here we characterize this effect. Recall the pixel-wise expressions for the exact and approximate filter, respectively: 

\begin{equation}
z_i = \sum_{j=1}^n w_{ij} \; y_j, \:\:\:\:\:\:\:\:\:\:\:\:\:\:\: \widehat{z}_i = \sum_{j=1}^n \widehat{w}_{ij} \; y_j
\end{equation}

\noindent The variance in the output pixel in terms of the variance in the input pixel is given by the sum-squared of the filter weights. That is,

\begin{equation}
\mbox{var}(z_i) =  \left( \sum_{j=1}^n w^2_{ij} \right)  \mbox{var}(y_i)  = \nu_i \; \mbox{var}(y_i)
\end{equation}
\begin{equation}
\mbox{var}(\widehat{z}_i) = \left( \sum_{j=1}^n \widehat{w}^2_{ij}\right)  \mbox{var}(y_i) = \widehat{\nu}_i \; \mbox{var}(y_i) 
\end{equation}

\begin{figure*}[ht]
\begin{center}
  \includegraphics*[viewport=20 1 700 480,scale=0.5]{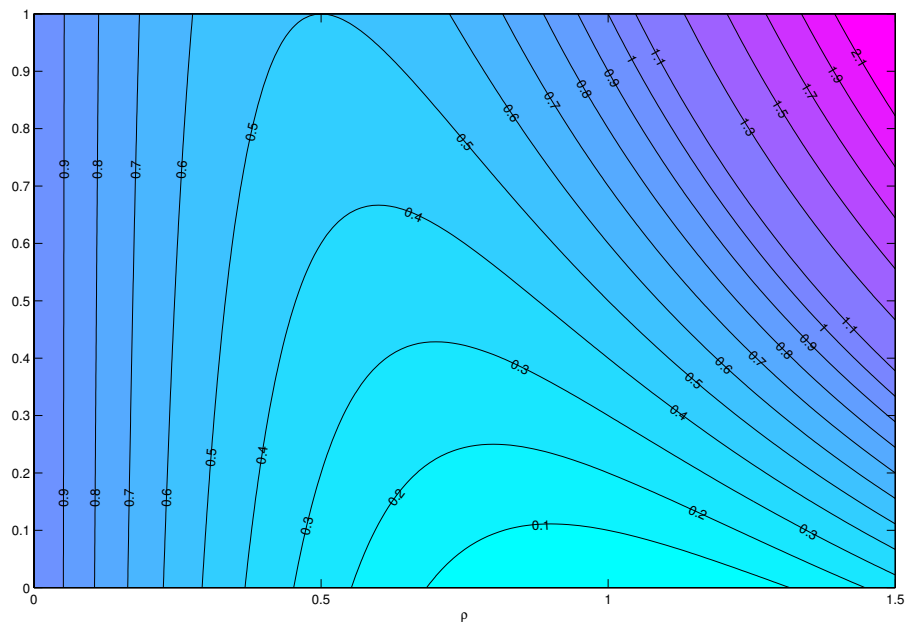}
\end{center}
\vspace{-6 mm}
 \caption{Values of $\widehat{\nu}^2$ as a function of $\rho$ (horizontal axis) and $\nu^2$ (vertical axis)}
\label{fig:sigsqcontours}
\end{figure*}

It is of interest to establish a relationship between the factors $\nu_i $ and $\widehat{\nu}_i$. We proceed as follows: 

\begin{eqnarray}
\widehat{\nu}_i & = & \widehat{\textbf{w}}_{i}^T \widehat{\textbf{w}}_{i}  \notag  \\
& = &  \left( \delta_i + \alpha^2 d_i^2 \left( \textbf{w}_i - \delta_i \right) \right)^T \left( \delta_i + \alpha^2 d_i^2 \left( \textbf{w}_i - \delta_i \right) \right) \notag \\
& = & \alpha^2 d_i^2 \nu_i +(\alpha^2 d_i^2 -2\alpha(1+\alpha)d_i + 1 + 2\alpha)
\end{eqnarray}
where $\delta_i$ is the shifted Dirac delta vector $[0,\cdots,0,1,0,\cdots,0]$, with subscript $i$ indicating that the value 1 occurs in the $i$-th position. The two variance factors are linearly related when $\alpha$ is small: 
\begin{equation}
\label{eq:sigmas}
\widehat{\nu}_i \approx   \rho_i^2 \: \nu_i + (\rho_i - 1)^2
\end{equation}
where $\rho_i = \alpha d_i$. The contour plot in Fig. \ref{fig:sigsqcontours}, shows the values of $\widehat{\nu}_i$ as a function of $\rho_i$ and $\nu_i$. Also, for the specific approximation pertaining to (\ref{finalroot}), we note that $d_i = \mathcal{O}(m)$ where $m$ is the size of the window over which filter weights are calculated. For instance, in the case of Fig. \ref{fig:Oldman}, $m = 11\times11 = 121$. Given $n$ pixels in the image, $\mbox{tr}(\mathbf{D}) = \mathcal{O}(mn)$. In the meantime, $\mbox{tr}(\mathbf{K}) = \mathcal{O}(n)$, $\mbox{tr}(\mathbf{KD^{-1}K}) = \mathcal{O}(n/m)$, $\mbox{tr}(\mathbf{K^2}) = \mathcal{O}(n^2)$, $\mbox{tr}(\mathbf{KD}) = \mathcal{O}(mn^2)$, and $\mbox{tr}(\mathbf{D^2}) = \mathcal{O}(m^2n^2)$. So for sufficiently large $m$ (typically larger than 5$\times$5), the terms $\mbox{tr}(\mathbf{D})$ and $\mbox{tr}(\mathbf{D^2})$ dominate the numerator and denominator as claimed.

\bibliographystyle{IEEEtran}   
\bibliography{refs}

\end{document}